\documentclass[sigconf]{acmart}
\usepackage{float}   
\usepackage{enumitem} 
\usepackage{makecell}  
\usepackage{subcaption} 
\usepackage{tikz} 
\usepackage{listings} 
\newcommand{\figref}[1]{\hyperref[#1]{Figure~\ref*{#1}}}
\newcommand{\tabref}[1]{\hyperref[#1]{Table~\ref*{#1}}}
\newcommand{\secref}[1]{\hyperref[#1]{Sec.~\ref*{#1}}}
\newcommand{\appref}[1]{\hyperref[#1]{App.~\ref*{#1}}}
\newcommand{\Appref}[1]{\hyperref[#1]{Appendix~\ref*{#1}}}
\AtBeginDocument{%
  }

\setcopyright{none}
\copyrightyear{2027}
\acmYear{2027}
\acmDOI{}
\acmConference[KDD '27]{Proceedings of the 33rd ACM SIGKDD Conference on
  Knowledge Discovery and Data Mining}{August 1--5, 2027}{San Jose, CA, USA}
\acmISBN{978-1-4503-XXXX-X/2027/08}

\begin{document}

\title[CADENCE: Interpretable ECG Foundation Models]{CADENCE: A Cardiac Atom Dictionary for Interpretable Neural Concept Extraction from ECG Foundation Models}

\author{Yixuan Duan}
\affiliation{%
  \department{Department of Electrical and Computer Engineering}
  \institution{Rice University}
  \city{Houston}
  \state{Texas}
  \country{USA}
}
\email{yd68@rice.edu}

\author{Arjun Naik}
\affiliation{%
  \department{Department of Computer Science}
  \institution{Rice University}
  \city{Houston}
  \state{Texas}
  \country{USA}
}
\email{an182@rice.edu}

\author{Sadeer Al-Kindi}
\affiliation{%
  \department{Department of Cardiology}
  \institution{Houston Methodist Hospital}
  \city{Houston}
  \state{Texas}
  \country{USA}
}

\email{sal-kindi@houstonmethodist.org}

\author{Wei Qiu}
\authornote{Corresponding author.}
\affiliation{%
  \department{Department of Electrical and Computer Engineering}
  \institution{Rice University}
  \city{Houston}
  \state{Texas}
  \country{USA}
}
\email{wq8@rice.edu}

\renewcommand{\shortauthors}{Duan et al.}

\begin{abstract}
 Foundation models for 12-lead electrocardiograms (ECGs) demonstrate strong transferability across clinical tasks, yet the physiological knowledge encoded in their representations remains opaque, limiting their interpretability, auditability, and scientific utility. We present \textbf{CADENCE} (\emph{Cardiac Atom Dictionary for Explainable Neural Concept Extraction}), a framework that decomposes the latent representations of an ECG foundation model into a human-interpretable and queryable dictionary of physiological concepts. Using a BatchTopK sparse autoencoder, CADENCE factorizes Layer-6 embeddings extracted from more than nine million ECG tokens into $8{,}192$ sparse cardiac \emph{atoms}. Compared with individual dimensions of the original dense embedding space, the learned atoms align substantially better with both clinical phenotypes and waveform morphology. CADENCE recovers a broad spectrum of cardiological concepts, including arrhythmias, conduction abnormalities, infarction and repolarization patterns, chamber and electrical-axis findings, and lead- and beat-phase-specific waveform primitives. At Layer 6, the best-performing atoms achieve mean AUROCs of $0.88$ for clinical phenotypes and $0.90$ for waveform morphology, compared with $0.78$ and $0.83$, respectively, for the best dense embedding dimensions. Sparse atom-based probes match or outperform dense embedding probes for phenotype, morphology, and age prediction while attributing each prediction to a small set of interpretable atoms. For phenotype prediction, the test AUROC improves from $0.93$ in the dense embedding space to $0.95$ in the sparse atom space. Moreover, atom-space geometry recovers physiologically coherent relationships, while targeted ablation of phenotype-associated atoms produces selective changes in frozen downstream model outputs. An automated large language model pipeline generates atom descriptions and quantitatively validates their descriptions by predicting held-out atom activations. Across independent external ECG datasets, CADENCE recovers overlapping physiological concepts and maintains consistent phenotype-prediction performance. Together, these findings establish CADENCE as a scalable framework for discovering and auditing the physiological knowledge encoded by ECG foundation models, providing a transparent foundation for scientific analysis and trustworthy clinical AI.
\end{abstract}

\ccsdesc[500]{Computing methodologies~Learning latent representations}
\ccsdesc[300]{Applied computing~Health informatics}
\ccsdesc[300]{Computing methodologies~Neural networks}
\ccsdesc[100]{Information systems~Data mining}

\keywords{Sparse autoencoders, Mechanistic interpretability,
  Foundation models, Electrocardiogram (ECG), Representation learning,
  Concept discovery, Interpretable AI in healthcare}


\maketitle

\section{Introduction}
\begin{figure*}[tb]
  \includegraphics[width=\textwidth]{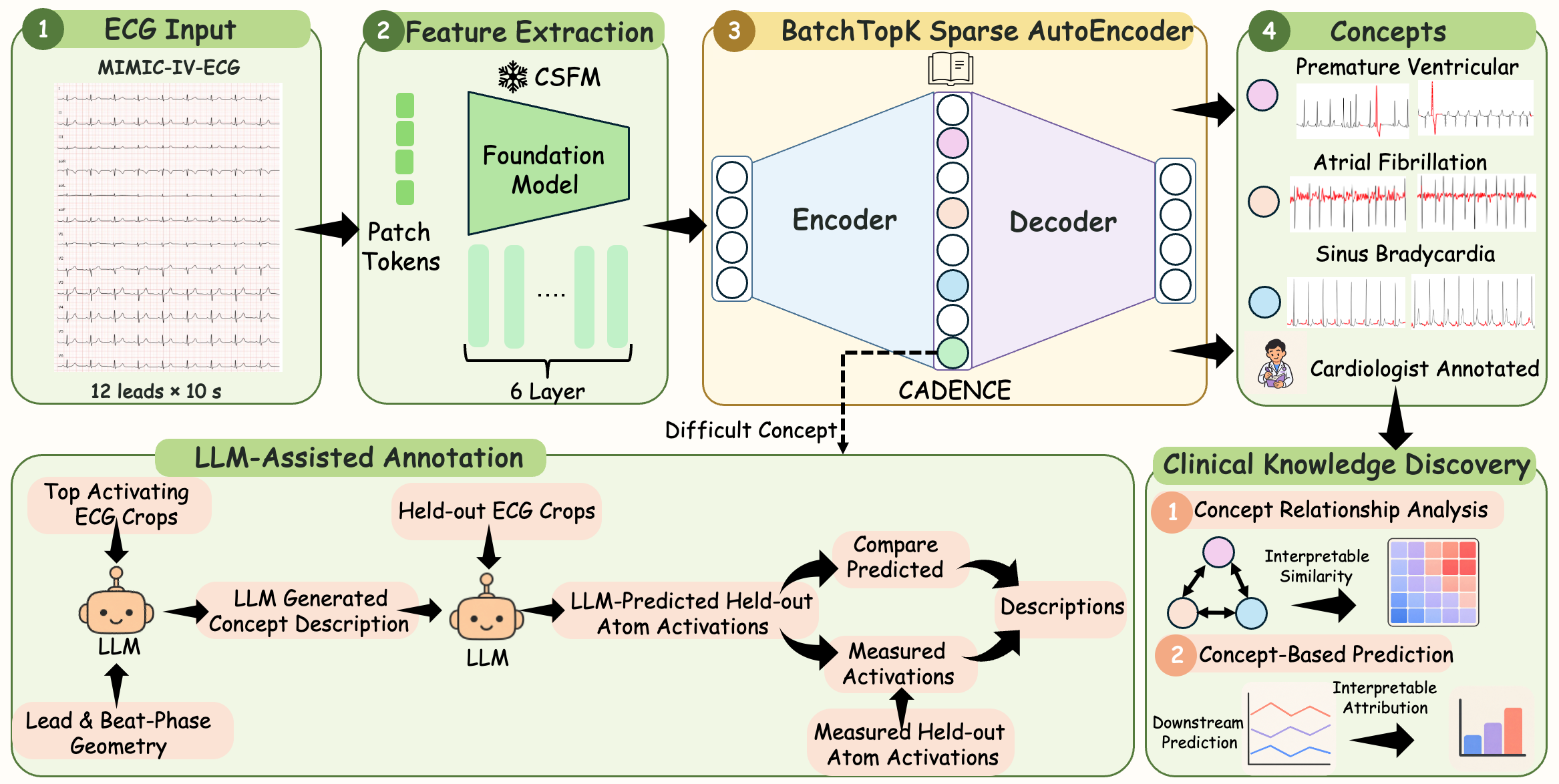}
  \caption{\textbf{Overview of CADENCE.} (1--3)~A frozen ECG foundation model
  (CSFM) encodes 12-lead MIMIC-IV-ECG recordings into patch tokens, which a
  BatchTopK sparse autoencoder decomposes into sparse cardiac atoms.
  (4)~Atoms are matched to clinical concepts and support concept relationship
  analysis and concept-based prediction. An LLM describes atoms from their
  top-activating ECGs and validates each description by predicting held-out
  activations.}
  \Description{Pipeline schematic. Top row, left to right: (1) a 12-lead,
  10-second MIMIC-IV-ECG recording; (2) a frozen six-layer CSFM foundation
  model producing patch tokens; (3) a BatchTopK sparse autoencoder whose
  encoder maps the representation to sparse cardiac atoms and whose decoder
  reconstructs it; (4) example atoms linked to premature ventricular
  complexes, atrial fibrillation, and sinus bradycardia with example ECG
  strips. Bottom right: clinical knowledge discovery via concept relationship
  analysis (a similarity heatmap) and concept-based prediction (attribution
  bar chart). Bottom left: an LLM describes difficult atoms from top-activating
  crops and lead and beat-phase geometry, then predicts held-out activations
  that are compared with measured activations to validate the description.}
  \label{fig:overview}
\end{figure*}

Electrocardiography (ECG) is the most widely used cardiac diagnostic test, yet reading one is itself a difficult, expertise-intensive skill: clinically decisive information is carried by subtle, millisecond-scale morphological details distributed across twelve simultaneous leads, competent interpretation takes years of training~\cite{ecgcompetency}, and even physicians interpret ECGs with substantial and variable error rates~\cite{ecgaccuracy}. Precisely because the raw signal is hard for humans to fully parse, self-supervised ECG foundation models (FMs) that learn general-purpose representations and transfer to dozens of downstream tasks—arrhythmia detection, risk prediction, and screening—often from a single frozen backbone \cite{ecgfm,stmem,ecgfmopen} are attractive, but they also make interpretability more important, not less: understanding what such a model encodes is valuable both for trusting and auditing its predictions and as a route to surfacing the physiological structure hidden in the signal. Yet these gains come from high-dimensional embeddings whose internal structure is opaque: a practitioner can see that a model predicts left bundle branch block, but not what physiological evidence the representation encodes. In a clinical setting this opacity is a barrier to trust, auditing, and scientific use—we cannot easily ask an ECG FM what it has learned about the heart.

Interpreting these models is hard because individual neurons are \emph{polysemantic}: a network can represent more features than it has dimensions, packing them into \emph{superposition}, so each neuron responds to several unrelated features rather than mapping cleanly onto one clinical concept. Sparse autoencoders (SAEs) can undo this superposition, decomposing activations into a large dictionary of sparse, approximately monosemantic features, and have exposed interpretable structure in language~\cite{sae_llm}, protein~\cite{interplm}, and, more recently, genomic and single-cell foundation models---small gene language models~\cite{sae_genelm}, single-cell transcriptomic models~\cite{sae_scfm}, and computational-pathology foundation models~\cite{sae_pathology}. Whether the same holds for physiological time-series FMs---and whether the recovered features map onto real cardiology---remains open.

We answer this for 12-lead ECG with CADENCE (\emph{Cardiac Atom Dictionary for Explainable Neural Concept Extraction}). As illustrated in \figref{fig:overview} (steps 1--3), CADENCE trains a BatchTopK SAE on the residual stream of an ECG foundation model (CSFM) and factorizes its 6 layers patch-token embeddings into a dictionary of ${\sim}8{,}000$ sparse atoms: the encoder decomposes each embedding into a few active atoms and the decoder reconstructs it, so the dictionary is faithful to the model yet far more interpretable than raw neurons. This turns a black-box backbone into a queryable substrate for knowledge discovery.

Building on this dictionary, CADENCE drives three complementary analyses
(\figref{fig:overview}, step 4 and clinical knowledge discovery). \emph{Concept discovery} associates each atom with
clinical phenotypes and waveform morphology, showing that atoms align with
cardiology far better than raw neurons. \emph{Relationship analysis} reads off
physiologically coherent relations between concepts directly from atom-space
geometry, without relational supervision. \emph{Concept-based prediction} uses
sparse atom probes that match or exceed the dense embedding for phenotype,
morphology, and age while attributing each prediction to a small set of named
atoms. An automated LLM pipeline (\figref{fig:overview}, LLM-assisted annotation) then describes each
atom from its top-activating ECG crops and activation geometry and then validates
the description by predicting held-out atom activations, yielding
human-readable, quantitatively checked annotations at scale. Finally, we show
that the frozen MIMIC-trained dictionary transfers to an independent,
cardiologist-labeled external cohort (PTB-XL) without retraining, recovering the
same physiological concepts and maintaining phenotype-prediction performance.
Together, these findings establish CADENCE as a scalable framework that turns an
opaque ECG foundation model into a transparent, queryable dictionary of
physiological concepts---discovered, related, predictive, automatically described,
and externally validated---providing a foundation for scientific discovery and
trustworthy clinical AI.

Our code is publicly available at \url{https://github.com/JayDuan123/2027-KDD-Ai4s-track-CADENCE}.

Our contributions are:
\begin{itemize}[leftmargin=1.3em,itemsep=2pt,topsep=2pt,parsep=0pt]
  \item To our knowledge we are the first to open up a cardiac physiological-signal
  foundation model: we show that its opaque, high-dimensional embeddings can be
  pulled out of superposition into a dictionary of sparse, approximately
  monosemantic features that align with clinical cardiac concepts far better than
  the model's raw neurons.
  \item We introduce CADENCE, which trains a BatchTopK sparse autoencoder
  on the residual stream of an ECG foundation model to factorize it into
  ${\sim}8{,}000$ sparse cardiac atoms that reconstruct the representation
  faithfully yet are far more interpretable than raw neurons.
  \item Across phenotype and morphology prediction, CADENCE probes match or exceed
  probes on the dense embedding while attributing each prediction to a small set of
  named atoms---atoms that an automated LLM pipeline describes and validates at
  scale.
  \item CADENCE turns an opaque ECG foundation model into an auditable dictionary
  of physiologically grounded concepts: practitioners can inspect what a
  representation encodes, decompose each prediction into named atoms, and surface
  confounded or shortcut features.
\end{itemize}

\section{Related Work}

\paragraph{ECG foundation models.}
Self-supervised pre-training on large unlabeled ECG corpora now yields
general-purpose encoders that transfer across arrhythmia detection, risk
stratification, and screening tasks~\cite{hannun2019,ribeiro2020,attia2019},
typically via masked-signal~\cite{mae2022} or
contrastive objectives~\cite{clocs2021,mehari2022} over 12-lead recordings~\cite{ecgfm,stmem,ecgfmopen}. These models
deliver strong downstream accuracy from a single frozen backbone, but they
are used as black boxes: their embeddings are high-dimensional and their
learned structure is not directly inspectable. Our work takes such a
model as given and asks a complementary question: how to \emph{read out}
the physiological concepts it has
already learned.

\paragraph{Sparse autoencoders and mechanistic interpretability.}
A central obstacle to interpreting neural representations is
\emph{superposition}: networks encode more features than they have
dimensions, so individual neurons are polysemantic and do not align with
human concepts~\cite{superposition}. Sparse autoencoders (SAEs) address
this by learning an overcomplete, sparsely activating dictionary that
reconstructs the model's activations~\cite{olshausen1996}, yielding features that are far more
monosemantic than neurons~\cite{sae_llm,cunningham2023sae,templeton2024}. Subsequent work has scaled SAEs
to frontier language models and improved the sparsity mechanism, including
top-$k$ and BatchTopK formulations that fix the active-code budget and
stabilize training~\cite{makhzani2013ksparse,gao2024scaling,bussmann2024batchtopk}, alongside
Gated and JumpReLU variants that reduce activation
shrinkage~\cite{rajamanoharan2024gated,rajamanoharan2024jumprelu}. We adopt a BatchTopK SAE but apply it,
for the first time to our knowledge, to a physiological time-series
foundation model rather than to language.

\paragraph{Interpreting scientific foundation models with SAEs.}
Beyond language, SAEs have recently been used to interpret scientific
sequence models. InterPLM trains SAEs on protein language model
embeddings and recovers thousands of interpretable features that map onto
binding sites, structural motifs, and functional domains~\cite{interplm}. Related work trains and evaluates SAEs on
protein language models to surface both universal and family-specific
features and to propose mechanistic hypotheses~\cite{saeprot}. Beyond
proteins, SAEs have begun to open up other biological foundation models:
single-cell transcriptomic models such as Geneformer and scGPT~\cite{sae_scfm},
gene language models~\cite{sae_genelm}, and computational-pathology image
models~\cite{sae_pathology}, in each case recovering biologically meaningful
features from otherwise opaque embeddings. These
studies establish SAEs as a route from representation to biological
insight for scientific foundation models; we extend this paradigm to
\emph{physiological signals}, where features additionally carry
lead-specific spatial and beat-phase temporal structure, and where a
generative decoder lets us relate atoms back to waveforms.

\paragraph{Interpretability for medical and ECG models.}
Prior interpretability for ECG classifiers has largely relied on
post-hoc saliency and attribution maps~\cite{gradcam,integratedgrad,shap,ecginterp},
linear probing of hidden layers~\cite{alain2016probes},
or concept-based and concept-bottleneck models~\cite{tcav,conceptbottleneck}. Saliency
highlights \emph{where} a specific prediction looks but not \emph{what}
reusable concepts a representation encodes; probing requires a predefined
label for every concept of interest and cannot discover unnamed features.
By decomposing the backbone into a concept dictionary, our approach is
unsupervised at the feature level, discovers sub-diagnostic waveform
primitives that have no standard label, and quantifies concept structure
directly in the model's own representation.
Taken together, these gaps motivate a concept-discovery framework specifically
designed for physiological foundation models.

\begin{figure*}[t]
  \centering
  \includegraphics[width=\textwidth]{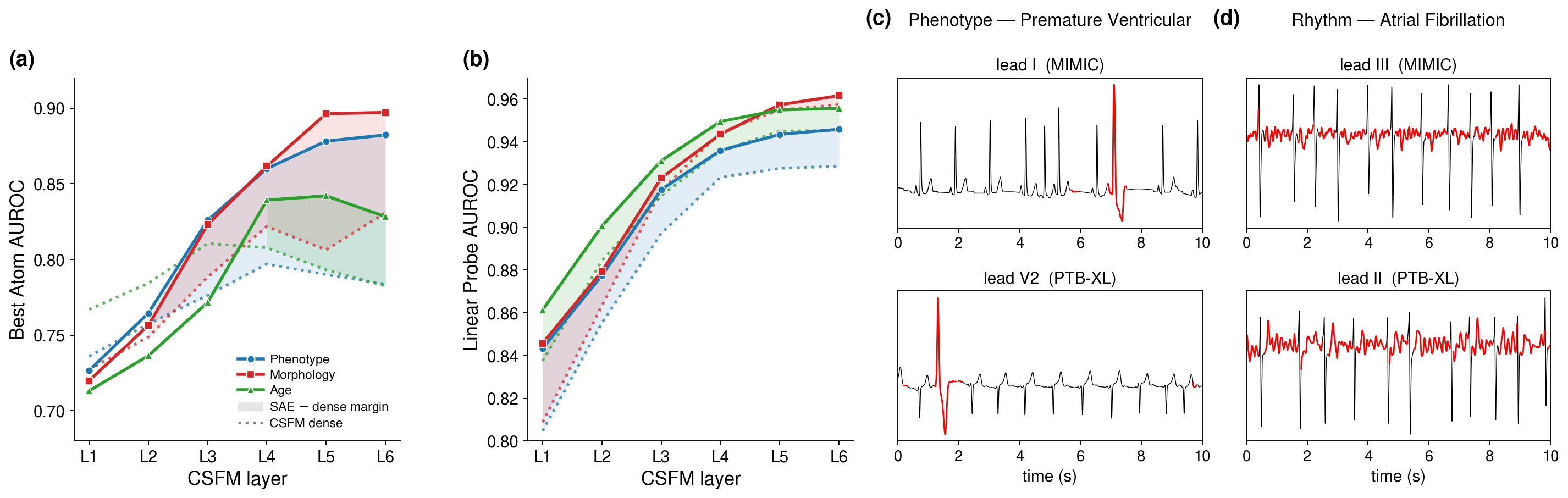}
  \caption{\textbf{What the dictionary encodes, how well it predicts, and
  where it fires.} \textbf{(a,b)}~Phenotype, morphology, and age across
  layers, solid \emph{SAE atoms} vs.\ dotted \emph{CSFM dense} with shaded
  SAE-over-dense margin: \textbf{(a)}~best single-atom AUROC,
  \textbf{(b)}~linear-probe test AUROC. \textbf{(c,d)}~One atom's red
  activated region on two 10\,s example ECGs, each shown on a MIMIC (top) and an
  external PTB-XL (bottom) record: \textbf{(c)}~a phenotype atom for PVC and
  \textbf{(d)}~a rhythm atom for atrial fibrillation firing on the fibrillatory
  baseline. A morphology (ST-segment) atom is in \figref{fig:st} and a
  rhythm atom in \figref{fig:rhythm}.}
  \label{fig:layers}
\end{figure*}

 \section{Methods}
\label{sec:methods}

\subsection{Datasets}
We build the dictionary on \textbf{MIMIC-IV-ECG}~\cite{mimicivecg,mimiciv,physionet},
a large public collection of de-identified 12-lead ECGs: a subject-disjoint
sample of $150{,}000$ ECGs for SAE training and a held-out ${\sim}15{,}000$ for
all downstream analyses (cohort details in \Appref{app:cohort}). For external
validation we apply the frozen, MIMIC-trained SAEs to \textbf{PTB-XL}~\cite{ptbxl},
an independent, cardiologist-labeled 12-lead cohort (\Appref{app:ptbxl}).

\subsection{ECG foundation model backbone}
We interpret a pre-trained 12-lead ECG foundation model (CSFM)~\cite{ecgfm}, a
transformer encoder~\cite{vaswani2017} with six layers and hidden width $d{=}768$. Each
recording is preprocessed identically to pre-training: signals are
resampled to 250\,Hz, cleaned per channel, and $z$-normalized per lead.
The model tokenizes each lead into non-overlapping 100\,ms patches
(25 samples), yielding $12\ \text{leads}\times100\ \text{patches}=1200$
patch tokens per ECG. We extract the residual-stream activations at
6 layers, $H\in\mathbb{R}^{1200\times d}$, as the target for dictionary
learning; the backbone is frozen throughout.

\subsection{Sparse autoencoder}
We factorize the 6 layers activations with a BatchTopK sparse autoencoder
(SAE)~\cite{gao2024scaling,bussmann2024batchtopk}. For a standardized
token $x\in\mathbb{R}^{d}$ (per-feature mean/scale estimated once on a
random token sample), the encoder produces latent codes and the decoder
reconstructs the input,
\begin{equation}
  z = \mathrm{BatchTopK}_k\!\big(W_{\mathrm{enc}}(x-b_{\mathrm{dec}})+b_{\mathrm{enc}}\big),
  \qquad
  \hat{x} = W_{\mathrm{dec}}\,z + b_{\mathrm{dec}}.
\end{equation}
where BatchTopK retains, on average, the $k$ largest activations per token
across the batch, giving an overcomplete dictionary of
$F$ \emph{atoms}. We choose BatchTopK for its fixed average activation budget---which lets the
number of active atoms vary across tokens---and its stable training on large
overcomplete dictionaries. We use
$F{=}8192$ and $k{=}128$, chosen by a grid search
over dictionary size and sparsity that trades off reconstruction explained
variance, dead-atom fraction, and interpretability yield (\appref{app:grid}).
The training objective is reconstruction error with an auxiliary term that
lets the most-dead latents explain the residual, mitigating dead atoms,
and the decoder columns are unit-normalized after each step:
\begin{equation}
  \mathcal{L} = \lVert x-\hat{x}\rVert_2^2
    + \alpha\,\lVert (x-\hat{x}) - \hat{x}_{\mathrm{aux}}\rVert_2^2 .
\end{equation}
The SAE is trained on 6 layers tokens sampled from $\sim$150k ECGs. Each
recording contributes multiple tokens balanced across leads (60 per
recording), for a training corpus of roughly $9\times10^{6}$ (9M)
tokens. We report reconstruction
explained variance, the average active-code count $L_0$, and the number of
dead atoms as dictionary-quality diagnostics.

\subsection{Atom--concept association and fingerprints}
To relate atoms to clinical concepts we use per-ECG atom activations,
defined as the maximum activation of each atom over the ECG's tokens.
We use the maximum activation because many ECG abnormalities are temporally
localized (e.g., PVCs, pacing spikes, ST changes), and averaging would dilute
these sparse events.
For a binary concept $c$ (e.g., a phenotype label) and atom $j$, the
association is the rank-based AUROC~\cite{hanley1982} of $a_j$ against $c$, folded to
$[0.5,1]$ to be sign-agnostic. The \emph{best-atom} score of a concept at a
layer is the largest such association over the dictionary, evaluated on the
held-out test set. We compare concepts by their best-atom AUROC---the single
most concept-selective atom---because interpretability hinges on whether a
concept is captured by one nameable unit. We summarize each concept by an
\emph{atom-space fingerprint}: the length-$F$ vector of signed AUROCs of
every atom against $c$. Concept--concept similarity is the cosine of these
fingerprints; as an activation-based alternative we also use a
\emph{prototype} fingerprint, the mean $z$-scored atom activation over the
concept's positive ECGs. We additionally derive continuous and
measurement-based concepts (heart rate, PR/QRS/QT intervals, electrical
axis, and thresholded morphology such as wide-QRS or prolonged QTc, using
standard diagnostic thresholds~\cite{kligfield2007}) from
the recordings' machine measurements. The complete set of concepts CADENCE
is evaluated against---diagnostic phenotypes, continuous measurements,
derived interval concepts, demographics, and morphology primitives, with
their definitions and cohort prevalences---is listed in
\tabref{tab:concepts}.

\subsection{Sparse probing for downstream prediction}

To test whether atoms are predictive as well as interpretable, we fit
$\ell_2$-regularized logistic-regression probes~\cite{sklearn2011} on the per-ECG atom
activations---a sparse-probing analysis over named features~\cite{gurnee2023}---and
compare them to probes on the dense 6 layers embedding,
using a subject-disjoint train/test split. Because atom activations are
sparse and named, each probe's prediction decomposes into a small set of
per-atom coefficient contributions, which we report as interpretable
attributions. For a given diagnosis we rank atoms strictly by probe
coefficient---with no manual reselection---and inspect the top few; atoms
that carry a large weight but rarely fire on the test set are kept and
explicitly marked as non-visualizable rather than dropped.

\subsection{Automated LLM interpretation and validation}
We scale interpretation from hand-reading to the dictionary with a
geometry-first LLM pipeline, following an InterPLM-style
describe-then-validate protocol in which a language model explains each
feature from its top activations~\cite{bills2023} (\figref{fig:overview}, LLM-assisted annotation).

We auto-interpret only atoms whose activation evidence can be constructed and
fairly validated, filtering out \emph{dead}, \emph{rare}, \emph{near-constant},
\emph{weak}, and \emph{lead-diffuse} atoms from pre-hoc activation and geometry
statistics. For each remaining atom the model produces a structured,
geometry-first description from its top-activating waveform crops and
activation-geometry summaries alone, with no machine-report text. We then
validate each description by having the model predict held-out activations from
the text alone, scored by a rank correlation; inclusion thresholds, prompting,
and validation details are in \Appref{app:autodesc}.

\section{Results}
\label{sec:results}

\subsection{Clinical concepts across network depth}

\label{sec:layers}

We first ask \emph{where} in the network different clinical concepts
become linearly accessible. For each layer $\ell\in\{1,\dots,6\}$ we train
a separate SAE, compute per-ECG feature activations, and score each concept by
its held-out best-atom AUROC (\secref{sec:methods}), taking, for the
per-feature comparison, the best value over concepts as each feature's
concept-association strength.

Across layers, SAE atoms are also far cleaner concept detectors than the raw
CSFM dimensions, reaching substantially higher per-feature association than the
dense embedding at every layer; we defer this atoms-vs-neurons comparison to
\Appref{app:layers} (\figref{fig:layers_ab}).

\paragraph{Concepts become more decodable with depth, and atoms beat the
dense embedding.} Grouping concepts into the three downstream categories used
later for prediction---phenotype (16 diagnoses), morphology (QRS/QTc/PR
abnormality), and age---and plotting the mean best-atom
AUROC per category against depth (\figref{fig:layers}a), two patterns
emerge. First, \textbf{every category becomes more decodable with depth}:
best-atom AUROC rises monotonically from Layer~1 to Layer~6 (phenotype
$0.73\!\to\!0.88$, morphology $0.72\!\to\!0.90$, age $0.71\!\to\!0.83$),
confirming that the backbone progressively concentrates clinical concepts
into individual directions the SAE can isolate. Second, \textbf{the best SAE
atom beats the best dense dimension, increasingly so with depth}: for
phenotype and morphology the SAE atom exceeds the CSFM dense embedding at
every layer (at Layer~6, phenotype $0.88$ vs.\ $0.78$, morphology $0.90$
vs.\ $0.83$), with the margin
widening from shallow to deep layers; the age signal, stronger in the dense
embedding early on, is matched and then exceeded by atoms by the deepest
layers. Together these results explain our choice
of Layer~6 for the main dictionary: clinical concepts are most linearly
separable, and most cleanly captured by single atoms, there.

\paragraph{Sparse atom probes match or exceed the dense embedding.}

\label{sec:probe}

We test whether imposing sparsity and interpretability costs predictive
accuracy by comparing, at each layer, a linear
probe fit on the dense CSFM embedding (mean-pooled, $768$-d) against one fit
on the SAE atom activations (max-pooled, $8192$-d), for three downstream task
families (\figref{fig:layers}b). Probes are elastic-net logistic
regressions trained on the training split and evaluated on a
subject-disjoint test set; for the phenotype and morphology panels we
report the mean test AUROC over the tasks in each family.

Across all three task families and all six layers, \textbf{the sparse atom
probe matches or exceeds the dense-embedding probe} (\figref{fig:layers}b)---interpretability
comes at no measurable accuracy cost. At the deepest layer it improves mean
test AUROC by $+0.017$ on phenotype ($0.929\!\to\!0.946$), $+0.011$ on age, and
$+0.005$ on morphology, a significant gain for phenotype (disjoint confidence
bands at every layer) and parity for morphology (\figref{fig:probe_ci}). The
advantage is largest in the \emph{shallow} layers---at Layer~1 the atom probe
reaches $0.843$ vs.\ $0.805$ on phenotype and $0.861$ vs.\ $0.837$ on
age---then narrows by Layers~5--6 as the backbone concentrates each concept
into more linearly separable directions, consistent with the SAE disentangling
features still superposed in the early-layer dense embedding. Because each atom
is named and each prediction decomposes into a handful of per-atom
contributions, the resulting probe is as accurate as the dense baseline yet
directly attributable.

\subsection{Clinical concept discovery}

\label{sec:localize}

Overlaying an atom's activation onto the raw waveform reveals not
only \emph{that} an atom is concept-associated but \emph{what} it detects
and \emph{where}. Doing so for atoms of different kinds shows that the
dictionary is organized hierarchically, spanning three levels of temporal
and spatial abstraction (\figref{fig:layers}); the full set of phenotype
and morphology features CADENCE recovers, grouped by category, is listed in
\tabref{tab:features}.

\paragraph{Rhythm level (multi-beat timing).}
A sinus-bradycardia atom (L6/3082, best-atom AUROC $0.98$) activates on the
\emph{elongated diastolic baseline
between beats}---the prolonged TP/RR interval (\figref{fig:rhythm},
\appref{app:rhythm}).
Across both example records it fires on every inter-beat segment in lead~II
and stays silent through the QRS--T complexes, so its ``feature'' is a
property of the \emph{spacing} of beats. This shows that atoms can encode temporal structure that only exists
\emph{across} several beats: the SAE represents heart rate as a
distributed, beat-locked baseline detector. A single named atom captures such multi-beat features directly, firing on the spacing between beats.

\paragraph{Phenotype level (a discrete pathological event).}
A premature-ventricular-complex atom (L6/5530, AUROC $0.90$) behaves as an
\emph{event detector}: within an otherwise regular sinus rhythm it fires on
a \emph{single} premature, wide, bizarre QRS and its following pause, while
remaining silent on all surrounding normal beats (\figref{fig:layers}c).
The atom thus isolates one clinically meaningful, temporally sparse
abnormality per record. The same event
morphology is picked out by the corresponding PVC atom on an external
PTB-XL record (\figref{fig:layers}c, bottom), so the localization holds
across datasets. A second, cross-dataset phenotype example is atrial
fibrillation: an AF atom (L6/5715) fires on the irregular fibrillatory
baseline between QRS complexes on both a MIMIC and an external PTB-XL record
(\figref{fig:layers}d). On the surface ECG, atrial fibrillation replaces
organized P waves with continuous low-amplitude fibrillatory activity and an
irregularly irregular ventricular response~\cite{kligfield2007}; the atom fires
on exactly this fibrillatory baseline in the inter-QRS segments, localizing the
arrhythmia's diagnostic signature on the trace. Further atom localizations---sinus bradycardia, atrial flutter, paced rhythm,
and prolonged PR---are shown in the appendix
(\figref{fig:rhythm}--\figref{fig:pr}).

\paragraph{Morphology level (a sub-beat waveform primitive).}
At the finest scale, atoms localize sub-beat waveform \emph{primitives}: a
specific segment of \emph{every} beat in a specific lead. They are
reusable geometric building blocks that recur on each cardiac cycle and from
which higher-level ischemia, repolarization, and conduction concepts are
composed. We show two such localizations---a prolonged PR segment and an
ST-segment primitive---in the appendix (\figref{fig:pr}, \figref{fig:st}).

Taken together, the three atoms span timescales from multi-beat rhythm,
through a single-beat pathological event, down to a sub-beat morphological
segment, and range from lead-agnostic to strictly lead-specific, and each one
is made visible the same way---by highlighting where the atom fires on the raw
ECG. The SAE
dictionary is therefore a layered
vocabulary, from waveform primitives up to rhythm, each element localizable
on the signal.

\subsection{Concept relationship analysis}
\label{sec:relations}

\begin{figure}[t]
  \hspace*{-8mm}\includegraphics[width=\linewidth]{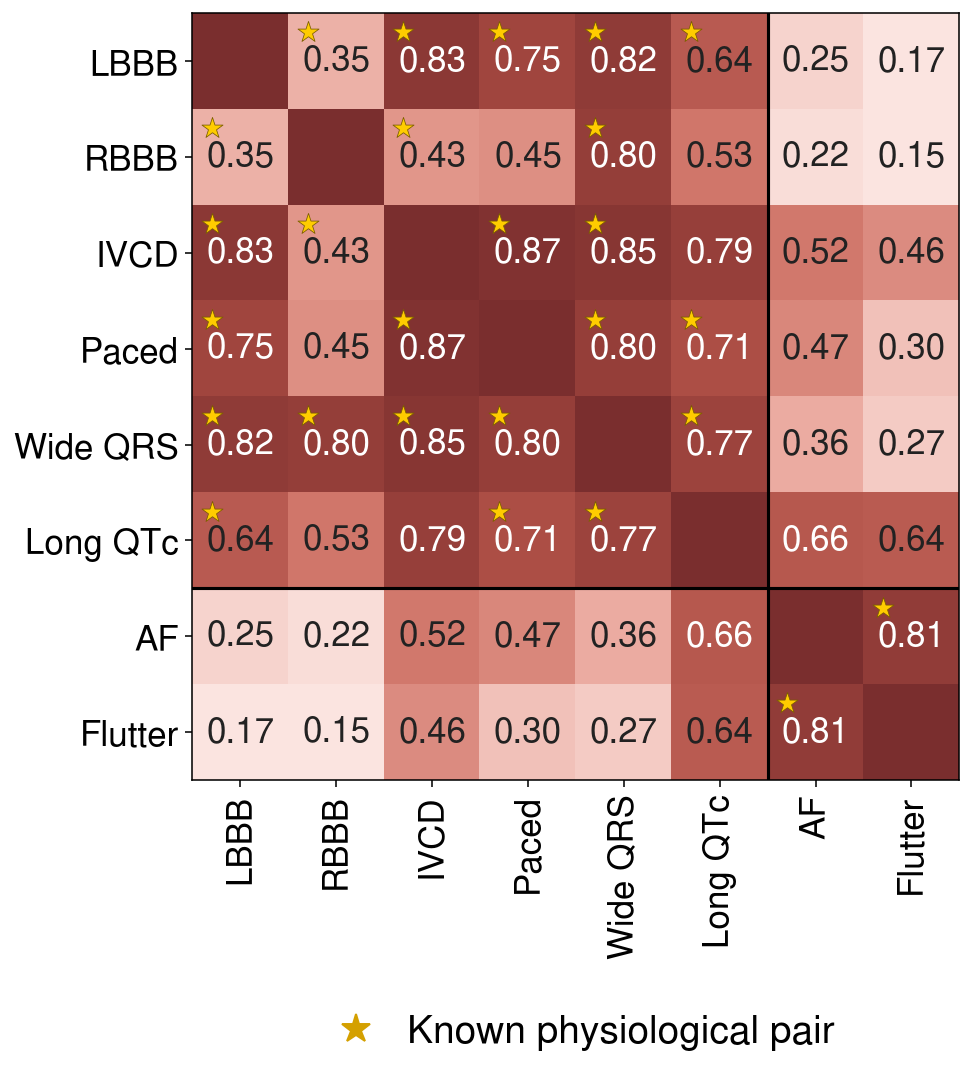}
  \vspace{-4mm}
  \caption{\textbf{Phenotype relationships in atom space.} Concept--concept
  cosine similarity of prototype fingerprints (yellow stars: known cardiology
  pairs).}
  \label{fig:relations}
\end{figure}

Beyond how well concepts are encoded, we ask whether the \emph{relations}
among concepts are recovered. We represent each concept by an atom-space
\emph{prototype} fingerprint---its mean $z$-scored atom activation over the
concept's positive ECGs---and measure concept--concept similarity as the
cosine of these fingerprints. Without any relational supervision, the curated
concepts self-organize into two physiologically coherent
blocks (\figref{fig:relations}). The first groups LBBB, RBBB,
IVCD, and paced rhythm together with the derived wide-QRS measurement
(pairwise cosine $0.80$--$0.87$), and places prolonged QTc adjacent to it
($0.64$--$0.79$)---recovering the interval-containment relationship that a
wider QRS mechanically lengthens the QT~\cite{kligfield2007,rautaharju2009}. Within this
block, LBBB and RBBB are the least similar pair ($0.35$),
consistent with their opposite ventricular activation~\cite{kligfield2007} despite a shared
wide QRS. A second block groups the atrial tachyarrhythmias---atrial
fibrillation and flutter (cosine $0.81$)---apart
from the conduction concepts, matching their distinct atrial origin~\cite{hindricks2020af}. The
recovered structure is robust to the fingerprint definition:
a signed-AUROC-fingerprint variant yields the same structure and agrees
essentially perfectly with the prototype cosine (\Appref{sec:appendix},
\figref{fig:relations_auroc}).

\subsection{Concept-based prediction}

\label{sec:sparseprobe}

\begin{figure*}[t]
  \centering
  \begin{minipage}[t]{0.50\textwidth}
    \centering
    \begin{tikzpicture}
      \node[anchor=north west,inner sep=0] (ima)
        {\includegraphics[height=6.85cm]{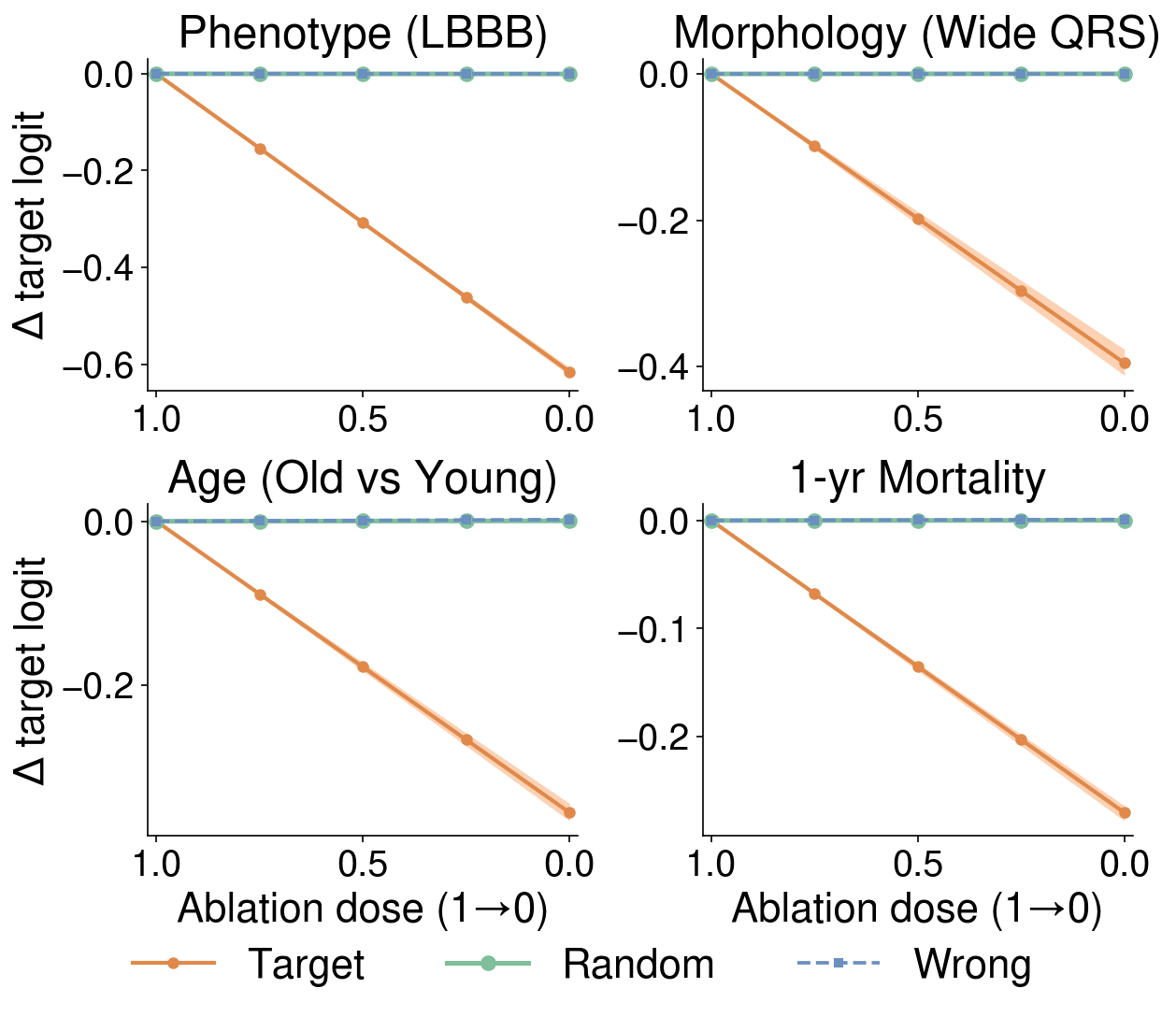}};
      \node[anchor=north west,inner sep=1pt,font=\bfseries] at (ima.north west) {(a)};
    \end{tikzpicture}
  \end{minipage}\hfill
  \begin{minipage}[t]{0.48\textwidth}
    \centering
    \begin{tikzpicture}
      \node[anchor=north west,inner sep=0] (imb)
        {\includegraphics[width=\linewidth]{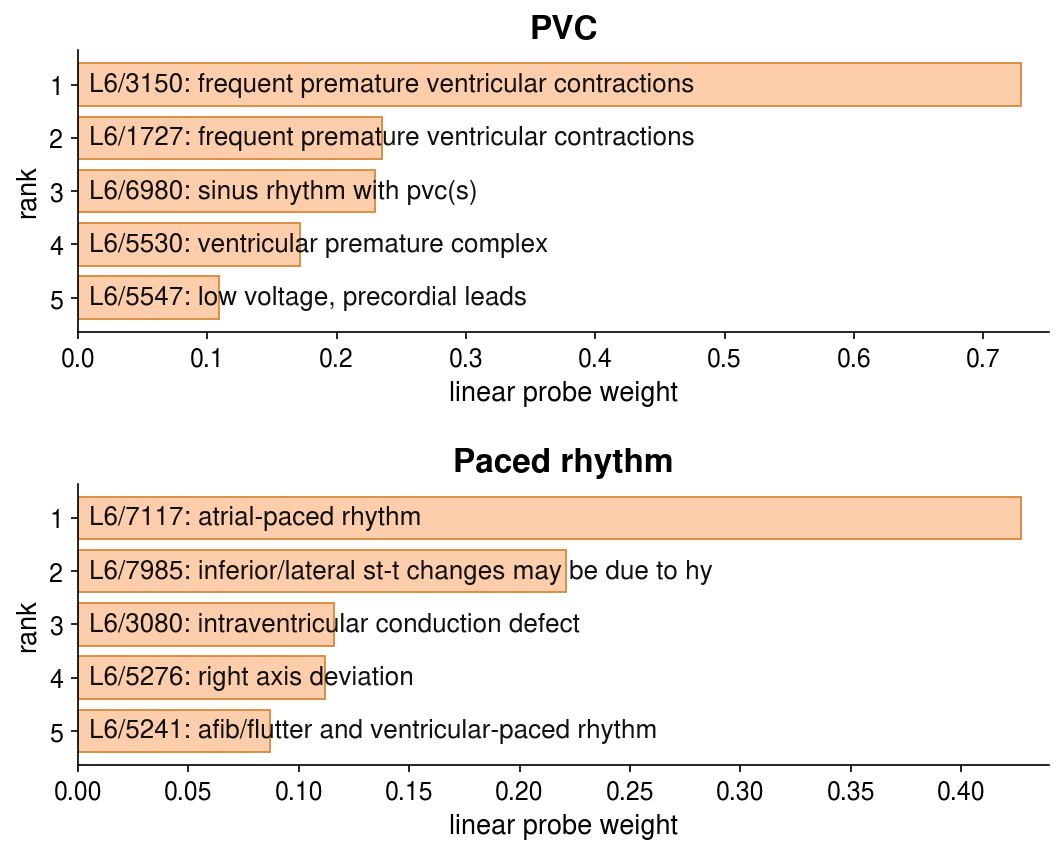}};
      \node[anchor=north west,inner sep=1pt,font=\bfseries] at (imb.north west) {(b)};
    \end{tikzpicture}
  \end{minipage}
  \caption{\textbf{Concept-level steering and sparse probing.} \textbf{(a)}~For four
  targets---LBBB, wide QRS, age, one-year mortality---the change in a frozen
  predictor's target logit as one Layer-6 atom is scaled from dose $1$ to $0$ on
  $N{=}80$ positives: the orange \emph{target} atom drives the logit down, while
  a frequency-matched green \emph{random} atom and a blue \emph{other-concept}
  atom do not; bands are 95\% bootstrap CIs. \textbf{(b)}~Sparse probing
  decomposes two diagnosis predictions---premature ventricular contractions
  (PVC, probe AUROC $0.947$; top) and paced rhythm ($0.951$; bottom)---into
  their five highest-weight atoms, each bar labeled by its most
  over-represented report statement (auxiliary).}
  \label{fig:relsteer}
\end{figure*}

\paragraph{Single atoms selectively steer the downstream readout.}

\label{sec:steer}

If atoms are genuinely the units in which the model represents a concept,
then removing a single concept-atom should selectively change that concept's
downstream prediction---an atom-level analogue of activation
steering~\cite{turner2023actadd,templeton2024}. We test this with an ablation audit
(\figref{fig:relsteer}a). For each target we train a frozen linear predictor
on the CSFM Layer-6 mean-pooled embedding---which never sees the SAE---and,
using an exact integrated-gradients attribution~\cite{integratedgrad} in atom space, identify the
single atom that most contributes to that target's logit. We then scale that
atom's activation from full to zero, decode the edited codes back to the
embedding, and read the frozen predictor, on $N=80$ high-confidence
target-positive ECGs. Two matched controls ablate instead a
frequency/magnitude-matched \emph{random} atom and a \emph{wrong} atom that
strongly encodes a different concept (atrial fibrillation) but is orthogonal
to the target readout.

Across all four targets, ablating the target atom drives the target logit
down monotonically with dose and by a large margin at full ablation
($-0.62$ for LBBB, $-0.40$ for wide-QRS, $-0.36$ for age, $-0.27$ for
mortality), whereas both the random and the wrong atom leave the target
logit unchanged. A single named atom
therefore acts as a selective control knob on the corresponding readout, for
targets as different as a conduction diagnosis, a QRS-width measurement,
patient age, and prognosis.

\paragraph{Sparse probing decomposes a diagnosis into named atoms.}

Sparse probing makes a diagnosis prediction legible by decomposing it into
a handful of atoms and showing, for each, the clinical statement it fires
on. We illustrate this for premature ventricular contractions (PVCs), where
the atom probe reaches test AUROC $0.947$ (\figref{fig:relsteer}b).
Strictly by probe weight, the top-five visualizable atoms are each
report-enriched for a ventricular-ectopy statement. The dominant-weight
atom (rank~1, coefficient $0.73$, roughly three times the next) is
$\sim\!90\times$ over-represented among ECGs the machine reads say have
``frequent premature ventricular contractions'' (lift $91.5$); ranks 2--4
carry the same frequent-PVC and ventricular-premature-complex statements.
Thus the probe's accuracy is not carried by an opaque combination of
features but by a small stack of atoms that each concentrate on exactly the
ECGs a clinician would call PVCs. Read out independently, the same atoms'
activation geometry is that of the ectopic beat---abnormal QRS-onset spikes
and secondary ST--T deflections (\secref{sec:llm})---so the report
enrichment says \emph{which} diagnosis and the geometry says \emph{what} on
the waveform.

The same decomposition holds for other diagnoses and is often anatomically
specific. For right bundle branch block (RBBB; probe AUROC $0.988$), the top
visualizable atoms are each report-enriched for (incomplete) RBBB and
rSR$'$(V1) statements, with lifts up to $\sim\!25\times$
(\figref{fig:probe_decomp_more}b). Their activation geometry localizes to the
terminal QRS and repolarization of the right precordial leads (V1/V2)---for
example a V2 terminal-QRS notch, the classic rSR$'$ signature---exactly the
diagnostic territory of RBBB. The same decomposition recurs across other
diagnoses---for example sinus bradycardia, whose probe
likewise concentrates on a handful of atoms whose enriched statements match the
target (\figref{fig:probe_decomp_more}a); the same procedure applies to the
morphology, age, and mortality probes (\Appref{sec:appendix}).

\subsection{LLM-assisted atom annotation}
\label{sec:llm}

\begin{figure*}[t]
  \centering
  \begin{minipage}[b]{0.40\textwidth}
    {\raggedright\textbf{(a)}\par}
    \centering
    \includegraphics[height=5.6cm]{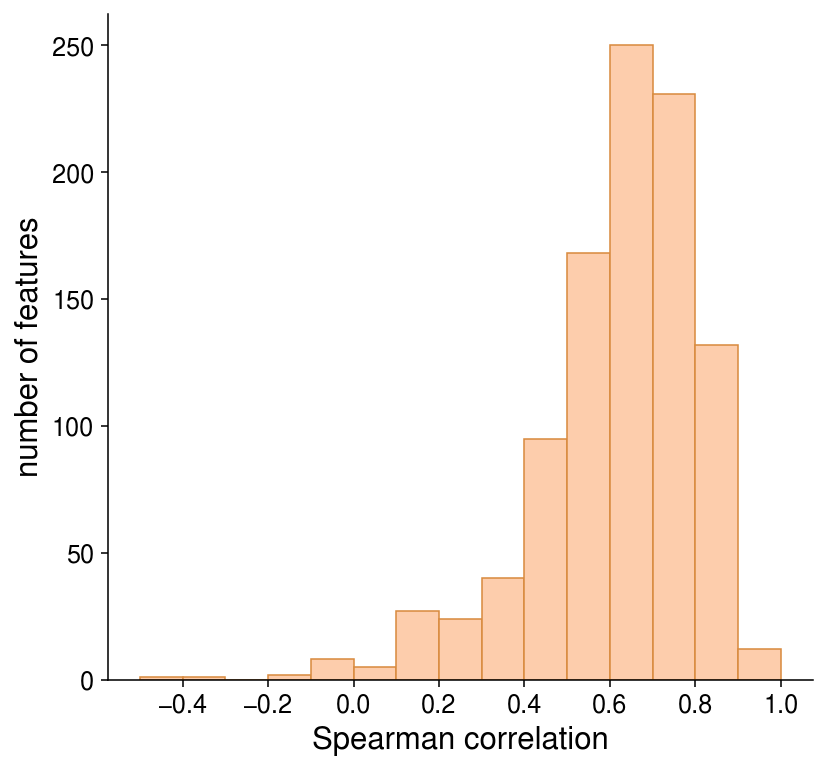}
  \end{minipage}\hspace{0.03\textwidth}%
  \begin{minipage}[b]{0.53\textwidth}
    {\raggedright\textbf{(b)}\par}
    \centering
    \includegraphics[height=5.6cm]{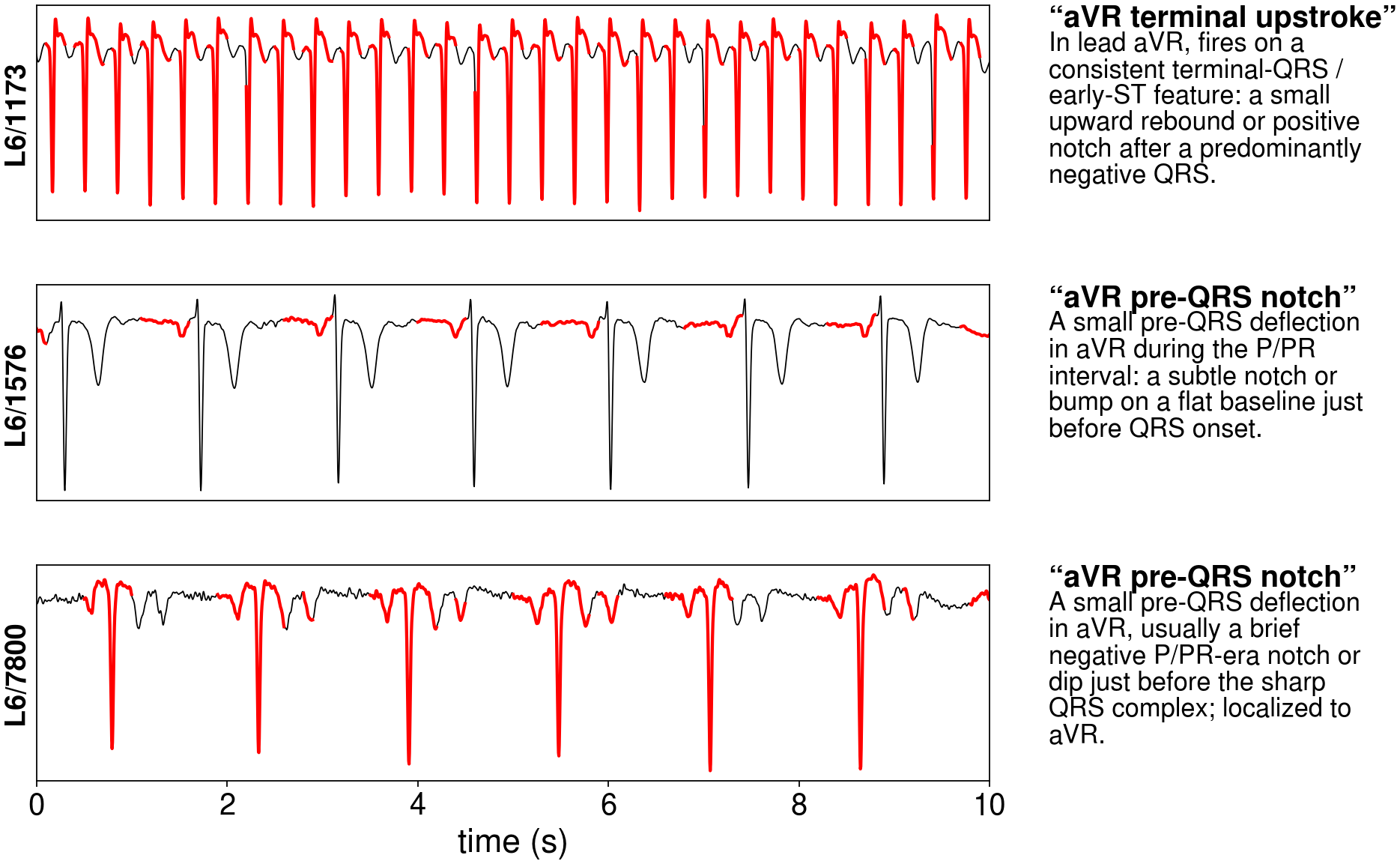}
  \end{minipage}
  \caption{\textbf{Automated LLM atom descriptions are quantitatively faithful.}
  \textbf{(a)}~Per-atom Spearman $\rho$ between the LLM's held-out activation
  predictions (from the text description alone) and measured activations, for
  $1000$ random atoms (gpt-5.4-mini); median $\rho=0.65$, with $92\%$ of atoms
  above $0.3$. \textbf{(b)}~Three example atoms: each atom's top-activating ECG
  (red marks where it fires) next to the model's geometry-based name and
  description.}
  \label{fig:llm}
\end{figure*}

We test whether the geometry-first descriptions of the SAE's concepts are
not merely plausible but \emph{quantitatively faithful}. Running the
describe-then-validate pipeline
(\figref{fig:overview}, LLM-assisted annotation) on a
random $1000$ of the $2991$ interpretable atoms (\figref{fig:llm}a), the
LLM's activation predictions---made from the text description alone, with
no access to activations---track the measured activation with a median
Spearman $\rho$ of $0.65$; $92\%$ of atoms
exceed $0.3$ and only $2\%$ fail ($\rho\le 0$).
Automated interpretation therefore extends from hand-read exemplars to the
dictionary at large: an LLM can read an atom's activation evidence and
write a description faithful enough to predict where that atom fires on
held-out ECGs, at a median rank correlation of $0.65$ over a thousand
atoms. As concrete examples, \figref{fig:llm}b pairs three atoms with their
top-activating ECG alongside the LLM's geometry-based name and description.

\section{Scientific and Clinical Contributions}
\label{sec:contributions}

\paragraph{Scientific contributions.}
CADENCE is, to our knowledge, the first application of sparse-autoencoder
dictionary learning to a \emph{physiological time-series} foundation model,
giving direct evidence that ECG FM embeddings are monosemantically
decomposable and store concepts in superposition---extending a phenomenon so
far documented mainly in language, vision, and protein models to a new,
clinically consequential modality. Beyond showing that such atoms \emph{exist},
we characterize \emph{how} an ECG backbone organizes them: clinical concepts
become progressively more linearly accessible with depth, and single atoms
increasingly outperform raw embedding dimensions. The recovered atoms span
three levels of temporal and spatial abstraction---multi-beat rhythm,
single-beat phenotype, and sub-beat morphology---so one dictionary captures
features from millisecond waveform primitives to whole-recording timing. Their
arrangement in atom space also reproduces physiologically coherent
relationships between concepts \emph{without any relational supervision},
indicating that clinical structure is written into the geometry of the
representation itself; this structure, and the concepts themselves, replicate
on an independent external cohort, indicating that they are properties of the
representation that generalize across datasets. Targeted ablation of a single named atom selectively changes
the corresponding frozen readout, showing atoms behave as functional units of
the model's representation, and an automated describe-then-validate pipeline
demonstrates that this interpretation can be scaled to the whole dictionary and
\emph{quantitatively} checked.

\paragraph{Clinical contributions.}
CADENCE turns an opaque ECG FM into an auditable, queryable dictionary of
physiologically grounded concepts, with several implications for how such
models could be used and trusted in cardiology. First, it makes a model's
evidence \emph{inspectable}: instead of accepting a prediction on faith, a
practitioner can ask which atoms it rests on and see the exact waveform
behavior each atom responds to---a premature-ventricular-complex atom firing on
the ectopic beat, or a right-bundle-branch-block atom on the terminal-QRS notch
in V1---evidence a cardiologist can check against their own reading. Second,
prediction becomes \emph{attributable}: sparse atom probes match or exceed the
dense embedding while decomposing each diagnosis into a small, named set of
contributing atoms, so accuracy no longer comes at the price of an opaque
score. Third, the same transparency surfaces \emph{failure modes}---confounded,
shortcut, or demographic features become visible for audit rather than hiding
inside a vector, a prerequisite for safe adoption. Fourth, because the
recovered concepts and their relationships replicate on an independent,
cardiologist-labeled external cohort without retraining, the dictionary behaves
as a cross-site auditing substrate, and
it can double as a hypothesis-generating tool---exposing which physiological
structure a model has actually learned about the heart.

\section{Conclusion}
\label{sec:conclusion}
We introduced CADENCE, which decomposes a frozen 12-lead ECG
foundation model into a sparse dictionary of physiological \emph{atoms} that
are far more concept-aligned than raw neurons. The dictionary recovers a
broad cardiology spectrum and reusable sub-diagnostic waveform primitives;
sparse atom probes are accurate and interpretable; atom-space geometry
reconstructs known physiological relationships; and an automated LLM pipeline
describes and validates every atom at scale, generalizing to an external
cohort. Together these establish sparse dictionary learning as a practical
route to auditing what physiological foundation models encode. Extending
CADENCE to other backbones, layers, and multi-center cohorts is a natural
next step toward trustworthy, auditable clinical AI.

\section{Limitations and Ethical Considerations}
\label{sec:ethics}

\paragraph{Limitations.}
Our study has several limitations. First, atom naming is bounded by current
cardiology knowledge and fixed geometric heuristics, so some interpretations may be
incomplete. Second, we focus on standard resting 12-lead ECGs; single-lead, ambulatory,
and longer continuous recordings are outside the present scope.

\paragraph{Ethical considerations.}
All experiments use publicly available, de-identified ECG data---the
MIMIC-IV-ECG resource~\cite{mimicivecg,mimiciv,physionet} for training and the
PTB-XL cohort~\cite{ptbxl} for external validation---accessed under their
respective PhysioNet data-use agreements. MIMIC-IV was reviewed by the Beth
Israel Deaconess Medical Center Institutional Review Board, which granted a
waiver of informed consent and approved the data-sharing initiative. For
PTB-XL, the relevant institutional ethics committee approved publication of the
anonymized data in an open-access database (PTB-2020-1). We add no identifiers,
do not attempt re-identification, and report only de-identified waveform
examples and aggregate, non-identifying analyses. The atoms are not diagnostic devices: not regulatorily cleared
and not for clinical decisions.

\section{Generative AI Usage}
\label{sec:genai}
Generative AI is used at two levels. As a studied component of the method,
our interpretation pipeline (\figref{fig:overview}, LLM-assisted annotation; \secref{sec:llm}) uses
large language models to describe and validate atoms. As an authoring aid, AI
assistants helped copy-edit and polish the writing; they were not used to
design the study, alter data or results, or draw conclusions. All AI-assisted outputs were reviewed and verified by the
authors, who take full responsibility for the content.

\bibliographystyle{ACM-Reference-Format}
\bibliography{references}

\appendix
\setcounter{figure}{0}\renewcommand{\thefigure}{A\arabic{figure}}
\setcounter{table}{0}\renewcommand{\thetable}{A\arabic{table}}

\clearpage   

\section{Additional Results}
\label{sec:appendix}

\subsection{Hyperparameter selection}
\label{app:grid}
We selected the dictionary size $F$ and sparsity $k$ by a grid search over
$F\!\in\!\{4096,8192,16384,32768\}$ and $k\!\in\!\{32,64,128,256\}$ at
Layer-6 (\figref{fig:grid}), scoring each configuration on three axes:
reconstruction explained variance (EV, higher is better), the fraction of
dead atoms (lower is better, since dead atoms waste dictionary capacity),
and the number of \emph{phenotype-specific} atoms---atoms that align
selectively with a single clinical phenotype---as a proxy for
interpretability yield (higher is better).

Three regularities decide the choice. (i)~\textbf{EV rises with $k$ but is
essentially flat in $F$}: at $k{=}128$, EV moves only $0.966\!\to\!0.969$ as
$F$ grows $8\times$, so a larger dictionary buys no reconstruction. (ii)~
\textbf{Dead atoms grow steeply with both $F$ and $k$}: at $k{=}128$ the dead
fraction jumps from $21\%$ ($F{=}8192$) to $49\%$ ($16384$) to $70\%$
($32768$), i.e., most of a larger dictionary is wasted. (iii)~\textbf{Interpretability
yield peaks at $k{=}128,\,F{=}8192$} ($459$ phenotype-specific atoms, the
global maximum of the grid); enlarging $F$ does not add specific atoms
($458$ at $16384$, $440$ at $32768$) while denser codes ($k{=}256$) sharply
reduce them. The chosen configuration ($k{=}128,\,F{=}8192$; orange star)
thus sits at the Pareto knee: EV within $0.01$ of the $k{=}256$ ceiling at
half the density, an acceptable $21\%$ dead fraction, and the maximum number
of phenotype-specific atoms. This is consistent with the corpus-size
saturation reported in \appref{app:cohort}: neither more tokens nor a
larger dictionary improves the decomposition at this backbone width.

\subsection{SAE atoms vs.\ neurons across depth}
\label{app:layers}
SAE atoms are far cleaner concept detectors than the raw CSFM embedding
dimensions, and the gap widens with depth (\figref{fig:layers_ab}).
Comparing the per-feature best held-out AUROC of SAE atoms against the dense
embedding dimensions and a random-noise control (\figref{fig:layers_ab}),
atoms reach substantially higher peak association than dense dimensions at
every layer, with the strongly-associated tail expanding with depth, while
the noise control stays at the $0.5$ floor.

\subsection{Downstream probes with bootstrap confidence intervals}
\label{app:probeci}
\begin{figure*}[tp]
  \centering
  \includegraphics[width=\textwidth]{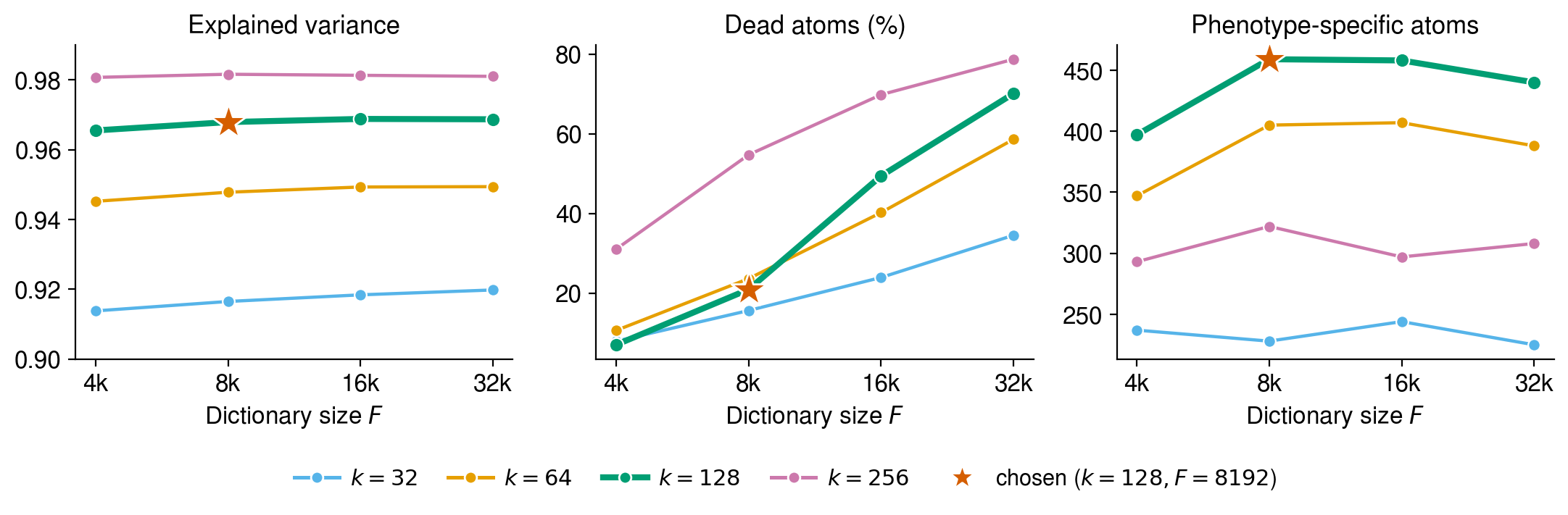}
  \caption{\textbf{SAE hyperparameter grid search at Layer-6.} Each panel
  plots one selection criterion against dictionary size $F$, with a line per
  sparsity level $k$. Left to right: explained variance rises with $k$ but is
  flat in $F$; dead-atom fraction grows steeply with both; and
  phenotype-specific atom count, our interpretability proxy, peaks at the
  chosen configuration $k{=}128,\,F{=}8192$, marked by the orange star.}
  \label{fig:grid}
  \Description{Three line-chart panels sharing an x-axis of dictionary size F
  (4k to 32k) with four lines for k in 32,64,128,256. Left: explained variance
  separates by k and is flat across F. Middle: dead-atom percentage increases
  with F and k. Right: phenotype-specific atom count peaks at k=128, F=8192,
  marked with a star.}

  \vspace{8pt}
  \includegraphics[width=0.58\textwidth]{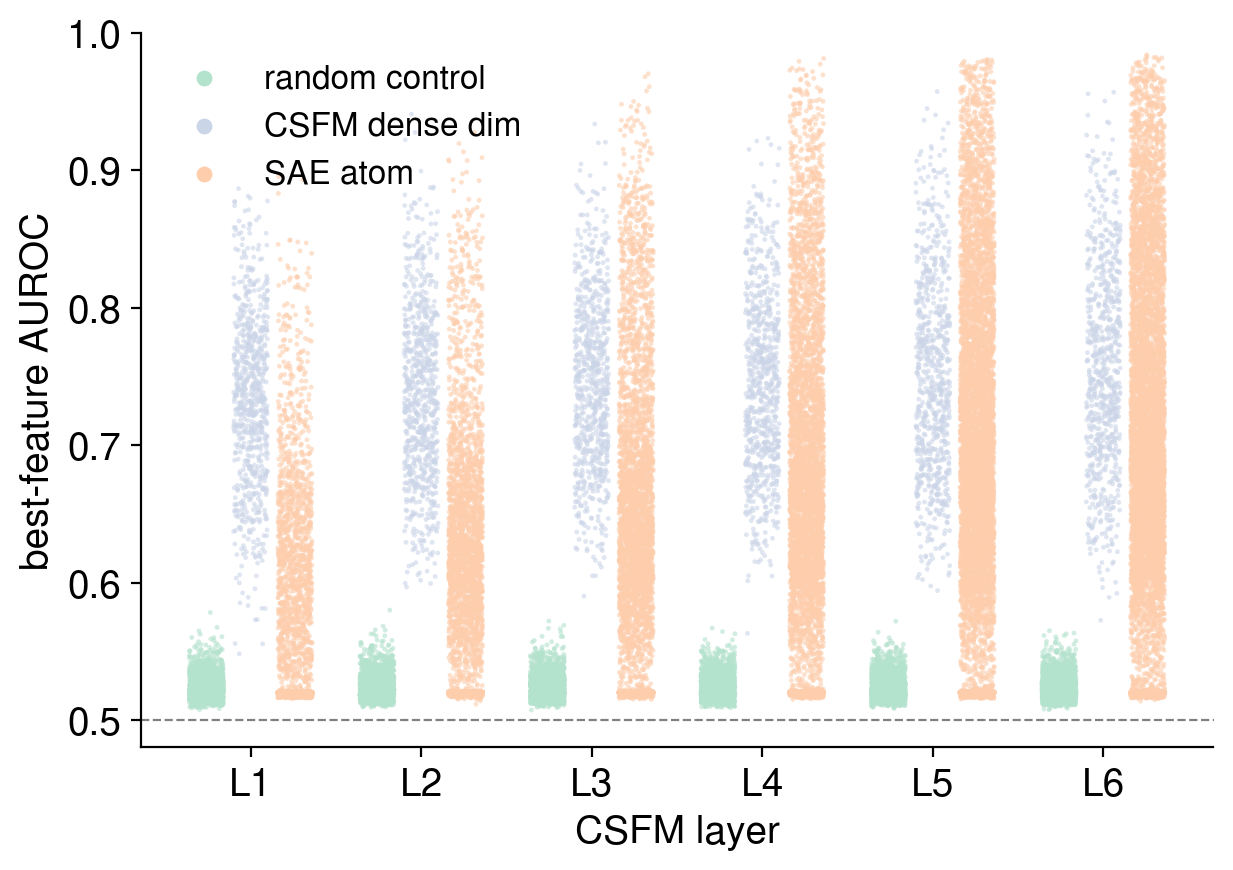}
  \caption{\textbf{SAE atoms are cleaner concept detectors than neurons.}
  Best held-out concept-association AUROC per feature at each layer, for a green
  random-noise control, blue CSFM dense embedding dimensions, and orange SAE
  atoms.}
  \label{fig:layers_ab}

  \vspace{8pt}
  \includegraphics[width=0.5\textwidth]{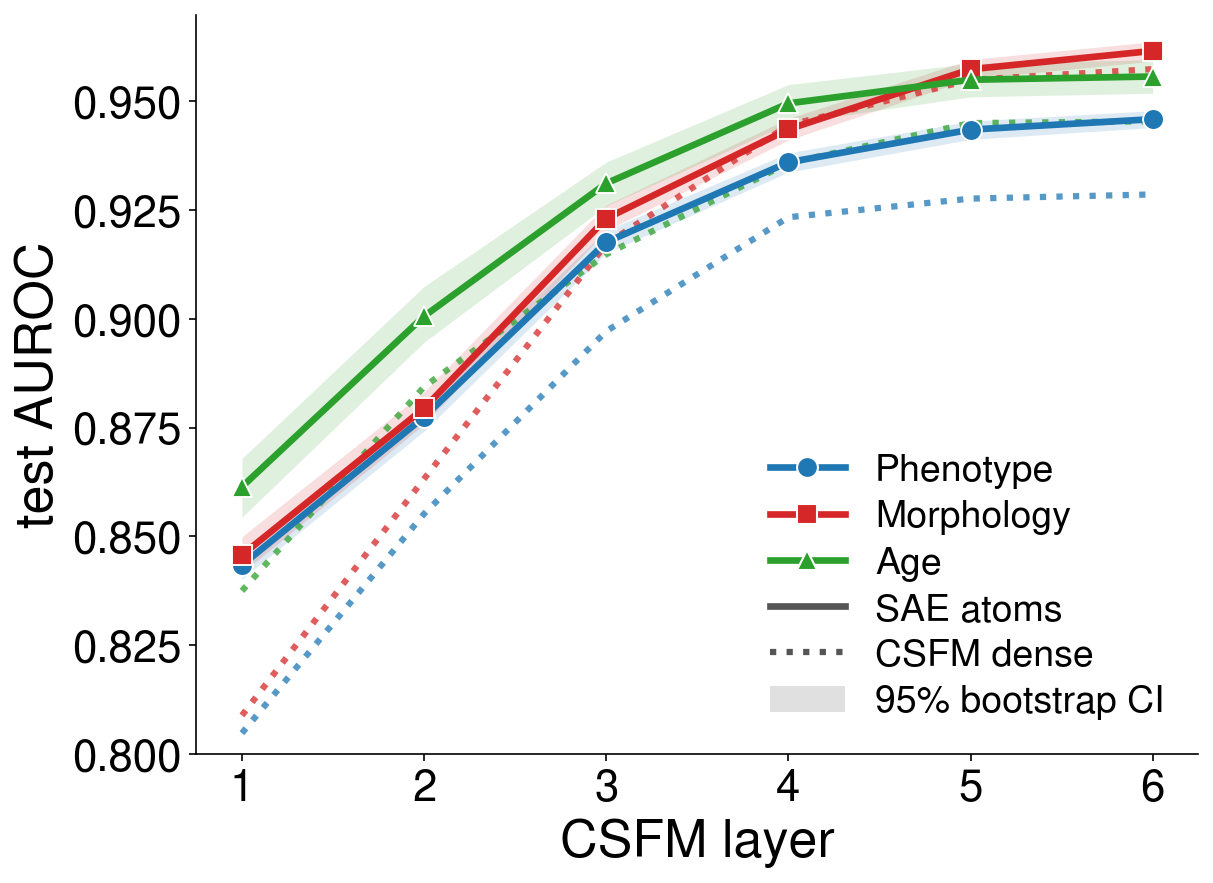}
  \caption{\textbf{Downstream probe performance with bootstrap CIs},
  cf.\ \figref{fig:layers}b. Per-layer test AUROC of the solid SAE-atom probe
  and the dotted CSFM dense embedding for the three task families; shaded
  bands are $95\%$ patient-level bootstrap confidence intervals for the SAE
  probe.}
  \label{fig:probe_ci}
\end{figure*}

\figref{fig:probe_ci} reproduces the per-layer probe comparison of
\figref{fig:layers}b with $95\%$ patient-level bootstrap confidence bands
around the SAE probe (in place of the SAE-over-dense margin shading). The CIs
are narrow at every layer, so the SAE-atom advantage over the dense embedding
is well outside sampling noise, most clearly in the shallow layers.

Table~\ref{tab:perconcept} breaks the aggregate best-atom curve of
\figref{fig:layers}a down to the individual concept level. For each concept
and layer it reports the best-atom AUROC of the SAE dictionary alongside the
best-unit AUROC of the raw CSFM dense embedding, each with a $95\%$
patient-level bootstrap confidence interval. The depth trend of
\figref{fig:layers}a holds concept by concept: the best single atom for most
concepts is found at L5--L6, and the sparse atom matches or exceeds the best
dense unit for almost every concept, with the margin widening in the deep
layers for conduction, rhythm, infarct and interval concepts.

\begin{figure*}[tp]
  \centering
  \includegraphics[width=\textwidth]{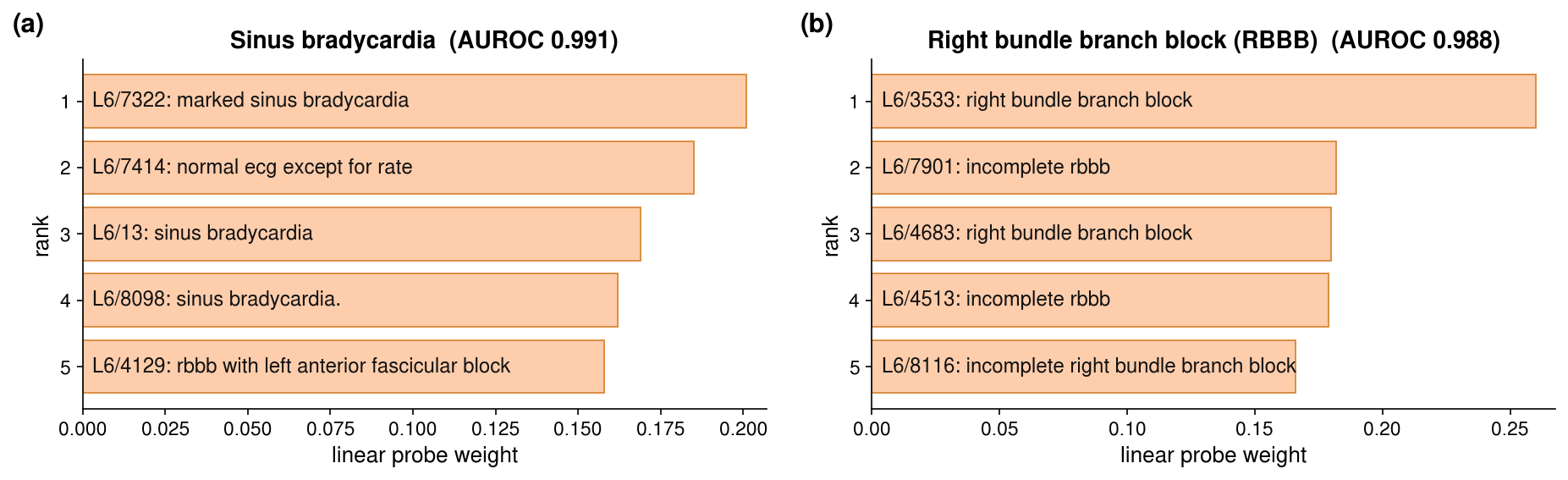}
  \caption{\textbf{Sparse-probe decomposition generalizes across diagnoses.}
  For two further phenotype probes, the five highest-weight atoms ranked by
  linear-probe weight, each labeled by its most over-represented (auxiliary)
  report statement: \textbf{(a)}~sinus bradycardia and \textbf{(b)}~right
  bundle branch block (RBBB) (test AUROC in each title). As for PVC and paced
  rhythm (\figref{fig:relsteer}b), each probe's weight concentrates on a
  small stack of atoms whose enriched statements match the target diagnosis;
  the RBBB atoms are each report-enriched for (incomplete) RBBB. Report
  enrichment shares its source with the labels and is auxiliary, not
  independent proof.}
  \label{fig:probe_decomp_more}
\end{figure*}

The sparse-probing decomposition of \figref{fig:relsteer}b is not specific to
PVC and paced rhythm. \figref{fig:probe_decomp_more} shows the same
decomposition for sinus bradycardia and right bundle branch block (RBBB): for
RBBB (panel~b; test AUROC $0.988$) the highest-weight atoms concentrate on
(incomplete) RBBB and rSR$'$(V1) report statements, and their activation
geometry localizes to the terminal QRS of the right precordial leads.

\begin{table*}[tp]
  \centering
  \caption{\textbf{Per-concept best-atom (SAE) vs.\ best-unit (CSFM dense) AUROC across the six CSFM layers, with 95\% patient-level bootstrap confidence intervals.}
  For each concept and layer we select, on the full test set, the single best sign-folded SAE atom and the single best sign-folded CSFM dense unit, then bootstrap that fixed unit's AUROC by resampling patients ($B{=}1000$). The larger of the two point estimates in each layer is in bold. Sparse atoms match or exceed the dense embedding for almost every concept, and the margin widens with depth for conduction, rhythm, infarct and interval concepts. This is the concept-level, dense-comparison decomposition of the aggregate best-atom curve in \figref{fig:layers}a.}
  \label{tab:perconcept}
  \footnotesize
  \setlength{\tabcolsep}{4pt}
  \renewcommand{\arraystretch}{1.0}
  \resizebox{\textwidth}{!}{%
  \begin{tabular}{l cc cc cc cc cc cc}
    \toprule
    & \multicolumn{2}{c}{L1} & \multicolumn{2}{c}{L2} & \multicolumn{2}{c}{L3} & \multicolumn{2}{c}{L4} & \multicolumn{2}{c}{L5} & \multicolumn{2}{c}{L6} \\
    \cmidrule(lr){2-3}\cmidrule(lr){4-5}\cmidrule(lr){6-7}\cmidrule(lr){8-9}\cmidrule(lr){10-11}\cmidrule(lr){12-13}
    Concept & SAE & dense & SAE & dense & SAE & dense & SAE & dense & SAE & dense & SAE & dense \\
    \midrule
    Atrial fibrillation & \makecell{0.677\\{\tiny [.66,.69]}} & \makecell{\textbf{0.724}\\{\tiny [.71,.74]}} & \makecell{0.716\\{\tiny [.70,.73]}} & \makecell{\textbf{0.724}\\{\tiny [.71,.74]}} & \makecell{\textbf{0.867}\\{\tiny [.86,.88]}} & \makecell{0.821\\{\tiny [.81,.83]}} & \makecell{\textbf{0.901}\\{\tiny [.89,.91]}} & \makecell{0.864\\{\tiny [.86,.87]}} & \makecell{\textbf{0.950}\\{\tiny [.94,.96]}} & \makecell{0.891\\{\tiny [.88,.90]}} & \makecell{\textbf{0.954}\\{\tiny [.95,.96]}} & \makecell{0.910\\{\tiny [.90,.92]}} \\
    Atrial flutter & \makecell{0.700\\{\tiny [.68,.72]}} & \makecell{\textbf{0.783}\\{\tiny [.77,.80]}} & \makecell{0.730\\{\tiny [.71,.75]}} & \makecell{\textbf{0.793}\\{\tiny [.78,.81]}} & \makecell{\textbf{0.823}\\{\tiny [.81,.84]}} & \makecell{0.820\\{\tiny [.81,.84]}} & \makecell{\textbf{0.844}\\{\tiny [.83,.86]}} & \makecell{0.837\\{\tiny [.82,.85]}} & \makecell{\textbf{0.872}\\{\tiny [.86,.89]}} & \makecell{0.846\\{\tiny [.83,.86]}} & \makecell{\textbf{0.874}\\{\tiny [.86,.89]}} & \makecell{0.854\\{\tiny [.84,.87]}} \\
    Sinus bradycardia & \makecell{0.723\\{\tiny [.71,.73]}} & \makecell{\textbf{0.843}\\{\tiny [.84,.85]}} & \makecell{0.758\\{\tiny [.75,.77]}} & \makecell{\textbf{0.869}\\{\tiny [.86,.87]}} & \makecell{\textbf{0.894}\\{\tiny [.89,.90]}} & \makecell{0.883\\{\tiny [.88,.89]}} & \makecell{\textbf{0.981}\\{\tiny [.98,.98]}} & \makecell{0.923\\{\tiny [.92,.93]}} & \makecell{\textbf{0.981}\\{\tiny [.98,.98]}} & \makecell{0.957\\{\tiny [.95,.96]}} & \makecell{\textbf{0.984}\\{\tiny [.98,.99]}} & \makecell{0.957\\{\tiny [.95,.96]}} \\
    PVC & \makecell{0.665\\{\tiny [.64,.68]}} & \makecell{\textbf{0.709}\\{\tiny [.69,.73]}} & \makecell{0.706\\{\tiny [.69,.73]}} & \makecell{\textbf{0.717}\\{\tiny [.70,.73]}} & \makecell{\textbf{0.767}\\{\tiny [.75,.78]}} & \makecell{0.738\\{\tiny [.72,.76]}} & \makecell{\textbf{0.853}\\{\tiny [.84,.87]}} & \makecell{0.747\\{\tiny [.73,.76]}} & \makecell{\textbf{0.905}\\{\tiny [.89,.91]}} & \makecell{0.779\\{\tiny [.76,.79]}} & \makecell{\textbf{0.904}\\{\tiny [.89,.91]}} & \makecell{0.805\\{\tiny [.79,.82]}} \\
    Paced rhythm & \makecell{0.753\\{\tiny [.72,.78]}} & \makecell{\textbf{0.822}\\{\tiny [.79,.85]}} & \makecell{0.803\\{\tiny [.77,.83]}} & \makecell{\textbf{0.845}\\{\tiny [.82,.87]}} & \makecell{0.834\\{\tiny [.80,.86]}} & \makecell{\textbf{0.859}\\{\tiny [.83,.88]}} & \makecell{\textbf{0.879}\\{\tiny [.85,.90]}} & \makecell{0.872\\{\tiny [.85,.89]}} & \makecell{\textbf{0.905}\\{\tiny [.88,.93]}} & \makecell{0.870\\{\tiny [.85,.89]}} & \makecell{\textbf{0.915}\\{\tiny [.89,.93]}} & \makecell{0.856\\{\tiny [.83,.88]}} \\
    LBBB & \makecell{\textbf{0.895}\\{\tiny [.88,.91]}} & \makecell{0.887\\{\tiny [.87,.90]}} & \makecell{0.928\\{\tiny [.92,.94]}} & \makecell{\textbf{0.941}\\{\tiny [.93,.95]}} & \makecell{\textbf{0.970}\\{\tiny [.96,.98]}} & \makecell{0.934\\{\tiny [.92,.94]}} & \makecell{\textbf{0.974}\\{\tiny [.97,.98]}} & \makecell{0.919\\{\tiny [.91,.93]}} & \makecell{\textbf{0.975}\\{\tiny [.97,.98]}} & \makecell{0.884\\{\tiny [.87,.89]}} & \makecell{\textbf{0.979}\\{\tiny [.97,.99]}} & \makecell{0.891\\{\tiny [.88,.90]}} \\
    RBBB & \makecell{\textbf{0.825}\\{\tiny [.81,.84]}} & \makecell{0.725\\{\tiny [.71,.74]}} & \makecell{\textbf{0.885}\\{\tiny [.88,.89]}} & \makecell{0.774\\{\tiny [.76,.79]}} & \makecell{\textbf{0.958}\\{\tiny [.95,.96]}} & \makecell{0.818\\{\tiny [.81,.83]}} & \makecell{\textbf{0.970}\\{\tiny [.96,.97]}} & \makecell{0.836\\{\tiny [.83,.84]}} & \makecell{\textbf{0.974}\\{\tiny [.97,.98]}} & \makecell{0.872\\{\tiny [.86,.88]}} & \makecell{\textbf{0.969}\\{\tiny [.96,.97]}} & \makecell{0.797\\{\tiny [.79,.81]}} \\
    LAFB & \makecell{\textbf{0.808}\\{\tiny [.80,.82]}} & \makecell{0.755\\{\tiny [.74,.77]}} & \makecell{\textbf{0.854}\\{\tiny [.84,.87]}} & \makecell{0.770\\{\tiny [.75,.79]}} & \makecell{\textbf{0.921}\\{\tiny [.91,.93]}} & \makecell{0.809\\{\tiny [.79,.82]}} & \makecell{\textbf{0.928}\\{\tiny [.92,.93]}} & \makecell{0.807\\{\tiny [.79,.82]}} & \makecell{\textbf{0.927}\\{\tiny [.92,.93]}} & \makecell{0.783\\{\tiny [.77,.80]}} & \makecell{\textbf{0.938}\\{\tiny [.93,.94]}} & \makecell{0.770\\{\tiny [.76,.78]}} \\
    Left-axis dev. & \makecell{\textbf{0.813}\\{\tiny [.81,.82]}} & \makecell{0.758\\{\tiny [.75,.77]}} & \makecell{\textbf{0.842}\\{\tiny [.83,.85]}} & \makecell{0.769\\{\tiny [.76,.78]}} & \makecell{\textbf{0.882}\\{\tiny [.87,.89]}} & \makecell{0.770\\{\tiny [.76,.78]}} & \makecell{\textbf{0.923}\\{\tiny [.92,.93]}} & \makecell{0.795\\{\tiny [.79,.81]}} & \makecell{\textbf{0.913}\\{\tiny [.91,.92]}} & \makecell{0.762\\{\tiny [.75,.77]}} & \makecell{\textbf{0.922}\\{\tiny [.92,.93]}} & \makecell{0.730\\{\tiny [.72,.74]}} \\
    LVH & \makecell{\textbf{0.656}\\{\tiny [.64,.67]}} & \makecell{0.653\\{\tiny [.64,.67]}} & \makecell{0.671\\{\tiny [.66,.68]}} & \makecell{\textbf{0.675}\\{\tiny [.66,.69]}} & \makecell{\textbf{0.682}\\{\tiny [.67,.70]}} & \makecell{0.678\\{\tiny [.67,.69]}} & \makecell{0.681\\{\tiny [.67,.70]}} & \makecell{\textbf{0.703}\\{\tiny [.69,.72]}} & \makecell{\textbf{0.667}\\{\tiny [.66,.68]}} & \makecell{0.659\\{\tiny [.65,.67]}} & \makecell{\textbf{0.684}\\{\tiny [.67,.70]}} & \makecell{0.631\\{\tiny [.62,.64]}} \\
    Low QRS voltage & \makecell{0.621\\{\tiny [.61,.63]}} & \makecell{\textbf{0.642}\\{\tiny [.63,.65]}} & \makecell{0.654\\{\tiny [.64,.67]}} & \makecell{\textbf{0.686}\\{\tiny [.67,.70]}} & \makecell{\textbf{0.673}\\{\tiny [.66,.68]}} & \makecell{0.668\\{\tiny [.66,.68]}} & \makecell{0.678\\{\tiny [.67,.69]}} & \makecell{\textbf{0.680}\\{\tiny [.67,.69]}} & \makecell{\textbf{0.704}\\{\tiny [.69,.72]}} & \makecell{0.664\\{\tiny [.65,.68]}} & \makecell{\textbf{0.712}\\{\tiny [.70,.72]}} & \makecell{0.654\\{\tiny [.64,.67]}} \\
    Inferior infarct & \makecell{\textbf{0.672}\\{\tiny [.66,.68]}} & \makecell{0.642\\{\tiny [.63,.65]}} & \makecell{\textbf{0.738}\\{\tiny [.73,.75]}} & \makecell{0.644\\{\tiny [.63,.66]}} & \makecell{\textbf{0.795}\\{\tiny [.79,.80]}} & \makecell{0.645\\{\tiny [.63,.66]}} & \makecell{\textbf{0.827}\\{\tiny [.82,.83]}} & \makecell{0.679\\{\tiny [.67,.69]}} & \makecell{\textbf{0.850}\\{\tiny [.84,.86]}} & \makecell{0.648\\{\tiny [.64,.66]}} & \makecell{\textbf{0.830}\\{\tiny [.82,.84]}} & \makecell{0.642\\{\tiny [.63,.65]}} \\
    Anterior infarct & \makecell{\textbf{0.756}\\{\tiny [.74,.77]}} & \makecell{0.711\\{\tiny [.70,.72]}} & \makecell{\textbf{0.737}\\{\tiny [.72,.75]}} & \makecell{0.707\\{\tiny [.69,.72]}} & \makecell{\textbf{0.830}\\{\tiny [.82,.84]}} & \makecell{0.715\\{\tiny [.70,.73]}} & \makecell{\textbf{0.831}\\{\tiny [.82,.84]}} & \makecell{0.702\\{\tiny [.69,.71]}} & \makecell{\textbf{0.828}\\{\tiny [.82,.84]}} & \makecell{0.686\\{\tiny [.67,.70]}} & \makecell{\textbf{0.836}\\{\tiny [.83,.85]}} & \makecell{0.647\\{\tiny [.63,.66]}} \\
    ST depression & \makecell{0.718\\{\tiny [.68,.75]}} & \makecell{\textbf{0.761}\\{\tiny [.73,.79]}} & \makecell{0.761\\{\tiny [.73,.79]}} & \makecell{\textbf{0.799}\\{\tiny [.77,.83]}} & \makecell{0.816\\{\tiny [.79,.84]}} & \makecell{\textbf{0.825}\\{\tiny [.80,.85]}} & \makecell{0.822\\{\tiny [.79,.84]}} & \makecell{\textbf{0.826}\\{\tiny [.80,.85]}} & \makecell{\textbf{0.810}\\{\tiny [.78,.83]}} & \makecell{0.786\\{\tiny [.76,.81]}} & \makecell{\textbf{0.822}\\{\tiny [.80,.84]}} & \makecell{0.790\\{\tiny [.77,.81]}} \\
    Prolonged PR & \makecell{\textbf{0.703}\\{\tiny [.69,.72]}} & \makecell{0.668\\{\tiny [.65,.68]}} & \makecell{\textbf{0.740}\\{\tiny [.73,.75]}} & \makecell{0.679\\{\tiny [.67,.69]}} & \makecell{\textbf{0.789}\\{\tiny [.78,.80]}} & \makecell{0.704\\{\tiny [.69,.72]}} & \makecell{\textbf{0.884}\\{\tiny [.87,.89]}} & \makecell{0.797\\{\tiny [.79,.81]}} & \makecell{\textbf{0.940}\\{\tiny [.93,.95]}} & \makecell{0.797\\{\tiny [.78,.81]}} & \makecell{\textbf{0.944}\\{\tiny [.94,.95]}} & \makecell{0.857\\{\tiny [.85,.87]}} \\
    Prolonged QT & \makecell{0.638\\{\tiny [.62,.65]}} & \makecell{\textbf{0.693}\\{\tiny [.68,.71]}} & \makecell{0.710\\{\tiny [.70,.72]}} & \makecell{\textbf{0.726}\\{\tiny [.71,.74]}} & \makecell{0.717\\{\tiny [.70,.73]}} & \makecell{\textbf{0.736}\\{\tiny [.72,.75]}} & \makecell{\textbf{0.785}\\{\tiny [.77,.80]}} & \makecell{0.766\\{\tiny [.75,.78]}} & \makecell{\textbf{0.851}\\{\tiny [.84,.86]}} & \makecell{0.756\\{\tiny [.74,.77]}} & \makecell{\textbf{0.850}\\{\tiny [.84,.86]}} & \makecell{0.746\\{\tiny [.73,.76]}} \\
    \bottomrule
  \end{tabular}}
\end{table*}

\begin{table*}[tp]
  \centering
  \caption{\textbf{Per-concept linear-probe AUROC (SAE atoms vs.\ CSFM dense embedding) across the six CSFM layers, with 95\% patient-level bootstrap confidence intervals.}
  For each concept and layer we fit an $\ell_1/\ell_2$-regularized logistic probe on the sparse atom vector (max-pooled) and, separately, on the dense CSFM embedding (mean-pooled), with a subject-held-out fit and a patient-level bootstrap ($B{=}500$) of the test AUROC. The larger of the two point estimates in each layer is in bold. The sparse atom probe matches or exceeds the dense probe for almost every concept and layer, with the largest gains in the shallow layers and for conduction, axis and infarct concepts; the two converge at the deep layers for the easiest rhythm concepts. This is the concept-level, per-layer decomposition of the aggregate probe comparison in \figref{fig:layers}b.}
  \label{tab:perconcept_probe}
  \footnotesize
  \setlength{\tabcolsep}{4pt}
  \renewcommand{\arraystretch}{1.0}
  \resizebox{\textwidth}{!}{%
  \begin{tabular}{l cc cc cc cc cc cc}
    \toprule
    & \multicolumn{2}{c}{L1} & \multicolumn{2}{c}{L2} & \multicolumn{2}{c}{L3} & \multicolumn{2}{c}{L4} & \multicolumn{2}{c}{L5} & \multicolumn{2}{c}{L6} \\
    \cmidrule(lr){2-3}\cmidrule(lr){4-5}\cmidrule(lr){6-7}\cmidrule(lr){8-9}\cmidrule(lr){10-11}\cmidrule(lr){12-13}
    Concept & SAE & dense & SAE & dense & SAE & dense & SAE & dense & SAE & dense & SAE & dense \\
    \midrule
    Atrial fibrillation & \makecell{\textbf{0.861}\\{\tiny [.85,.87]}} & \makecell{0.819\\{\tiny [.81,.83]}} & \makecell{\textbf{0.899}\\{\tiny [.89,.91]}} & \makecell{0.874\\{\tiny [.87,.88]}} & \makecell{\textbf{0.935}\\{\tiny [.93,.94]}} & \makecell{0.934\\{\tiny [.93,.94]}} & \makecell{\textbf{0.962}\\{\tiny [.96,.97]}} & \makecell{0.960\\{\tiny [.96,.96]}} & \makecell{\textbf{0.971}\\{\tiny [.97,.97]}} & \makecell{0.970\\{\tiny [.97,.97]}} & \makecell{\textbf{0.974}\\{\tiny [.97,.98]}} & \makecell{0.973\\{\tiny [.97,.98]}} \\
    Atrial flutter & \makecell{\textbf{0.854}\\{\tiny [.84,.87]}} & \makecell{0.834\\{\tiny [.82,.85]}} & \makecell{\textbf{0.871}\\{\tiny [.86,.88]}} & \makecell{0.866\\{\tiny [.85,.88]}} & \makecell{\textbf{0.897}\\{\tiny [.89,.91]}} & \makecell{0.894\\{\tiny [.88,.90]}} & \makecell{\textbf{0.906}\\{\tiny [.89,.92]}} & \makecell{0.905\\{\tiny [.89,.92]}} & \makecell{\textbf{0.914}\\{\tiny [.90,.92]}} & \makecell{0.911\\{\tiny [.90,.92]}} & \makecell{\textbf{0.916}\\{\tiny [.91,.93]}} & \makecell{0.913\\{\tiny [.90,.92]}} \\
    Sinus bradycardia & \makecell{\textbf{0.918}\\{\tiny [.91,.92]}} & \makecell{0.887\\{\tiny [.88,.89]}} & \makecell{\textbf{0.936}\\{\tiny [.93,.94]}} & \makecell{0.933\\{\tiny [.93,.94]}} & \makecell{\textbf{0.973}\\{\tiny [.97,.98]}} & \makecell{0.970\\{\tiny [.97,.97]}} & \makecell{\textbf{0.987}\\{\tiny [.98,.99]}} & \makecell{0.986\\{\tiny [.98,.99]}} & \makecell{\textbf{0.990}\\{\tiny [.99,.99]}} & \makecell{0.990\\{\tiny [.99,.99]}} & \makecell{\textbf{0.991}\\{\tiny [.99,.99]}} & \makecell{0.990\\{\tiny [.99,.99]}} \\
    PVC & \makecell{\textbf{0.765}\\{\tiny [.75,.78]}} & \makecell{0.743\\{\tiny [.73,.76]}} & \makecell{\textbf{0.792}\\{\tiny [.78,.81]}} & \makecell{0.764\\{\tiny [.75,.78]}} & \makecell{\textbf{0.880}\\{\tiny [.87,.89]}} & \makecell{0.808\\{\tiny [.79,.82]}} & \makecell{\textbf{0.925}\\{\tiny [.92,.93]}} & \makecell{0.870\\{\tiny [.86,.88]}} & \makecell{\textbf{0.942}\\{\tiny [.93,.95]}} & \makecell{0.902\\{\tiny [.89,.91]}} & \makecell{\textbf{0.946}\\{\tiny [.94,.95]}} & \makecell{0.915\\{\tiny [.91,.92]}} \\
    Paced rhythm & \makecell{\textbf{0.883}\\{\tiny [.86,.91]}} & \makecell{0.875\\{\tiny [.85,.90]}} & \makecell{\textbf{0.908}\\{\tiny [.89,.93]}} & \makecell{0.902\\{\tiny [.88,.92]}} & \makecell{\textbf{0.927}\\{\tiny [.91,.94]}} & \makecell{0.922\\{\tiny [.90,.94]}} & \makecell{0.938\\{\tiny [.92,.95]}} & \makecell{\textbf{0.938}\\{\tiny [.92,.96]}} & \makecell{\textbf{0.947}\\{\tiny [.93,.96]}} & \makecell{0.936\\{\tiny [.92,.95]}} & \makecell{\textbf{0.952}\\{\tiny [.94,.97]}} & \makecell{0.940\\{\tiny [.92,.96]}} \\
    LBBB & \makecell{\textbf{0.975}\\{\tiny [.97,.98]}} & \makecell{0.962\\{\tiny [.95,.97]}} & \makecell{\textbf{0.981}\\{\tiny [.97,.99]}} & \makecell{0.975\\{\tiny [.97,.98]}} & \makecell{\textbf{0.988}\\{\tiny [.98,.99]}} & \makecell{0.982\\{\tiny [.98,.99]}} & \makecell{\textbf{0.990}\\{\tiny [.99,.99]}} & \makecell{0.986\\{\tiny [.98,.99]}} & \makecell{\textbf{0.991}\\{\tiny [.99,.99]}} & \makecell{0.986\\{\tiny [.98,.99]}} & \makecell{\textbf{0.990}\\{\tiny [.99,.99]}} & \makecell{0.984\\{\tiny [.98,.99]}} \\
    RBBB & \makecell{\textbf{0.949}\\{\tiny [.94,.95]}} & \makecell{0.879\\{\tiny [.87,.89]}} & \makecell{\textbf{0.975}\\{\tiny [.97,.98]}} & \makecell{0.949\\{\tiny [.94,.95]}} & \makecell{\textbf{0.984}\\{\tiny [.98,.99]}} & \makecell{0.968\\{\tiny [.96,.97]}} & \makecell{\textbf{0.987}\\{\tiny [.98,.99]}} & \makecell{0.982\\{\tiny [.98,.98]}} & \makecell{\textbf{0.987}\\{\tiny [.99,.99]}} & \makecell{0.982\\{\tiny [.98,.98]}} & \makecell{\textbf{0.988}\\{\tiny [.99,.99]}} & \makecell{0.980\\{\tiny [.98,.98]}} \\
    LAFB & \makecell{\textbf{0.892}\\{\tiny [.88,.90]}} & \makecell{0.849\\{\tiny [.84,.86]}} & \makecell{\textbf{0.941}\\{\tiny [.94,.95]}} & \makecell{0.909\\{\tiny [.90,.92]}} & \makecell{\textbf{0.964}\\{\tiny [.96,.97]}} & \makecell{0.931\\{\tiny [.93,.94]}} & \makecell{\textbf{0.970}\\{\tiny [.97,.97]}} & \makecell{0.943\\{\tiny [.94,.95]}} & \makecell{\textbf{0.971}\\{\tiny [.97,.97]}} & \makecell{0.943\\{\tiny [.94,.95]}} & \makecell{\textbf{0.971}\\{\tiny [.97,.97]}} & \makecell{0.942\\{\tiny [.94,.95]}} \\
    Left-axis dev. & \makecell{\textbf{0.879}\\{\tiny [.87,.89]}} & \makecell{0.839\\{\tiny [.83,.85]}} & \makecell{\textbf{0.926}\\{\tiny [.92,.93]}} & \makecell{0.888\\{\tiny [.88,.89]}} & \makecell{\textbf{0.951}\\{\tiny [.95,.95]}} & \makecell{0.910\\{\tiny [.90,.92]}} & \makecell{\textbf{0.957}\\{\tiny [.95,.96]}} & \makecell{0.923\\{\tiny [.92,.93]}} & \makecell{\textbf{0.960}\\{\tiny [.96,.96]}} & \makecell{0.923\\{\tiny [.92,.93]}} & \makecell{\textbf{0.958}\\{\tiny [.96,.96]}} & \makecell{0.923\\{\tiny [.92,.93]}} \\
    LVH & \makecell{\textbf{0.746}\\{\tiny [.74,.76]}} & \makecell{0.715\\{\tiny [.70,.73]}} & \makecell{\textbf{0.790}\\{\tiny [.78,.80]}} & \makecell{0.759\\{\tiny [.75,.77]}} & \makecell{\textbf{0.839}\\{\tiny [.83,.85]}} & \makecell{0.821\\{\tiny [.81,.83]}} & \makecell{\textbf{0.865}\\{\tiny [.86,.87]}} & \makecell{0.849\\{\tiny [.84,.86]}} & \makecell{\textbf{0.872}\\{\tiny [.86,.88]}} & \makecell{0.851\\{\tiny [.84,.86]}} & \makecell{\textbf{0.875}\\{\tiny [.87,.88]}} & \makecell{0.853\\{\tiny [.84,.86]}} \\
    Low QRS voltage & \makecell{\textbf{0.721}\\{\tiny [.71,.73]}} & \makecell{0.694\\{\tiny [.68,.71]}} & \makecell{\textbf{0.768}\\{\tiny [.76,.78]}} & \makecell{0.763\\{\tiny [.75,.77]}} & \makecell{\textbf{0.832}\\{\tiny [.82,.84]}} & \makecell{0.827\\{\tiny [.82,.83]}} & \makecell{0.863\\{\tiny [.86,.87]}} & \makecell{\textbf{0.867}\\{\tiny [.86,.87]}} & \makecell{0.872\\{\tiny [.87,.88]}} & \makecell{\textbf{0.873}\\{\tiny [.87,.88]}} & \makecell{\textbf{0.874}\\{\tiny [.87,.88]}} & \makecell{0.870\\{\tiny [.86,.88]}} \\
    Inferior infarct & \makecell{\textbf{0.769}\\{\tiny [.76,.78]}} & \makecell{0.676\\{\tiny [.66,.69]}} & \makecell{\textbf{0.828}\\{\tiny [.82,.84]}} & \makecell{0.751\\{\tiny [.74,.76]}} & \makecell{\textbf{0.900}\\{\tiny [.89,.91]}} & \makecell{0.839\\{\tiny [.83,.85]}} & \makecell{\textbf{0.927}\\{\tiny [.92,.93]}} & \makecell{0.890\\{\tiny [.88,.90]}} & \makecell{\textbf{0.939}\\{\tiny [.93,.94]}} & \makecell{0.896\\{\tiny [.89,.90]}} & \makecell{\textbf{0.943}\\{\tiny [.94,.95]}} & \makecell{0.896\\{\tiny [.89,.90]}} \\
    Anterior infarct & \makecell{\textbf{0.834}\\{\tiny [.83,.84]}} & \makecell{0.772\\{\tiny [.76,.78]}} & \makecell{\textbf{0.859}\\{\tiny [.85,.87]}} & \makecell{0.828\\{\tiny [.82,.84]}} & \makecell{\textbf{0.890}\\{\tiny [.88,.90]}} & \makecell{0.855\\{\tiny [.85,.86]}} & \makecell{\textbf{0.911}\\{\tiny [.90,.92]}} & \makecell{0.891\\{\tiny [.88,.90]}} & \makecell{\textbf{0.919}\\{\tiny [.91,.93]}} & \makecell{0.885\\{\tiny [.88,.89]}} & \makecell{\textbf{0.919}\\{\tiny [.91,.93]}} & \makecell{0.877\\{\tiny [.87,.89]}} \\
    ST depression & \makecell{\textbf{0.848}\\{\tiny [.82,.87]}} & \makecell{0.817\\{\tiny [.79,.84]}} & \makecell{0.866\\{\tiny [.84,.89]}} & \makecell{\textbf{0.869}\\{\tiny [.85,.89]}} & \makecell{\textbf{0.893}\\{\tiny [.87,.91]}} & \makecell{0.892\\{\tiny [.87,.91]}} & \makecell{\textbf{0.916}\\{\tiny [.90,.93]}} & \makecell{0.912\\{\tiny [.90,.93]}} & \makecell{\textbf{0.914}\\{\tiny [.90,.93]}} & \makecell{0.902\\{\tiny [.88,.92]}} & \makecell{\textbf{0.926}\\{\tiny [.91,.94]}} & \makecell{0.909\\{\tiny [.89,.92]}} \\
    Prolonged PR & \makecell{\textbf{0.811}\\{\tiny [.80,.82]}} & \makecell{0.747\\{\tiny [.74,.76]}} & \makecell{\textbf{0.866}\\{\tiny [.86,.87]}} & \makecell{0.828\\{\tiny [.82,.84]}} & \makecell{\textbf{0.932}\\{\tiny [.93,.94]}} & \makecell{0.909\\{\tiny [.90,.92]}} & \makecell{\textbf{0.950}\\{\tiny [.94,.96]}} & \makecell{0.947\\{\tiny [.94,.95]}} & \makecell{\textbf{0.960}\\{\tiny [.95,.97]}} & \makecell{0.956\\{\tiny [.95,.96]}} & \makecell{\textbf{0.962}\\{\tiny [.96,.97]}} & \makecell{0.957\\{\tiny [.95,.96]}} \\
    Prolonged QT & \makecell{\textbf{0.789}\\{\tiny [.78,.80]}} & \makecell{0.767\\{\tiny [.75,.78]}} & \makecell{\textbf{0.832}\\{\tiny [.82,.84]}} & \makecell{0.826\\{\tiny [.82,.84]}} & \makecell{\textbf{0.897}\\{\tiny [.89,.90]}} & \makecell{0.893\\{\tiny [.89,.90]}} & \makecell{0.924\\{\tiny [.92,.93]}} & \makecell{\textbf{0.926}\\{\tiny [.92,.93]}} & \makecell{\textbf{0.947}\\{\tiny [.94,.95]}} & \makecell{0.936\\{\tiny [.93,.94]}} & \makecell{\textbf{0.950}\\{\tiny [.94,.95]}} & \makecell{0.937\\{\tiny [.93,.94]}} \\
    \bottomrule
  \end{tabular}}
\end{table*}

\subsection{Rhythm-level localization example}
\label{app:rhythm}
\begin{figure*}[tp]
  \centering
  \includegraphics[width=0.8\textwidth]{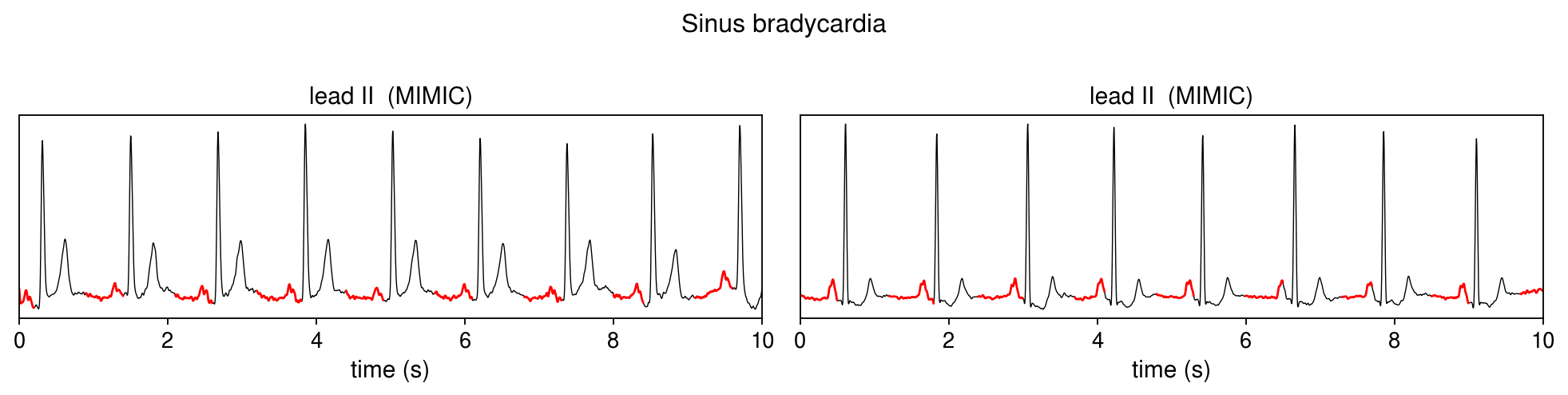}
  \caption{\textbf{Rhythm-level localization.} Sinus-bradycardia atom
  L6/3082 fires, shown red, on the elongated diastolic baseline between beats
  in lead~II across two 10\,s example ECGs, remaining silent through the
  QRS--T complexes.}
  \label{fig:rhythm}

  \vspace{8pt}
  \includegraphics[width=0.8\textwidth]{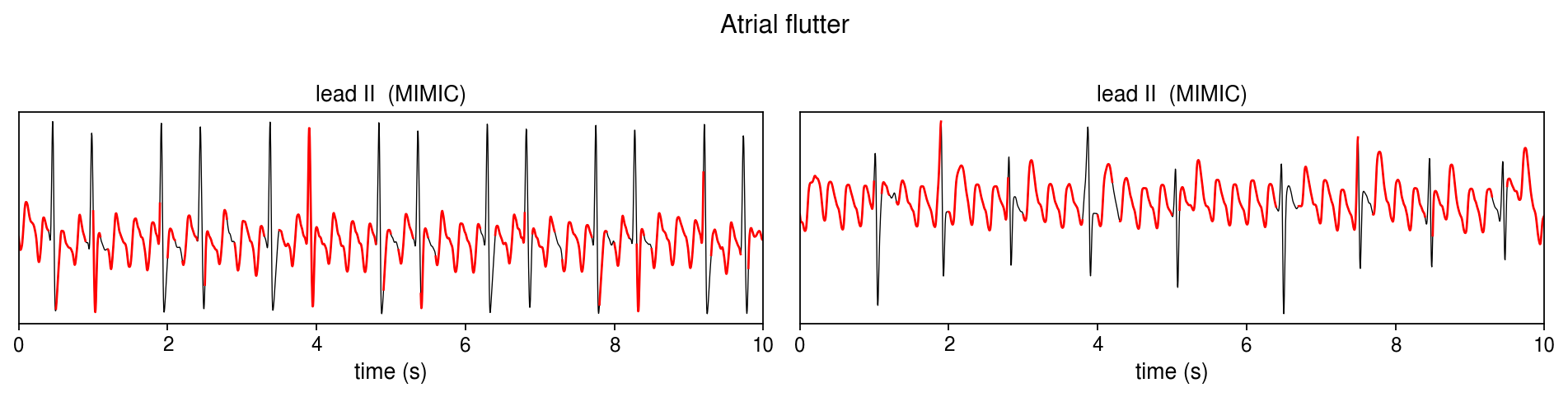}
  \caption{\textbf{Rhythm-level localization: atrial flutter.} Atom L6/418
  fires, shown red, on the regular sawtooth flutter waves between QRS
  complexes in lead~II across two 10\,s example ECGs.}
  \label{fig:flutter}

  \vspace{8pt}
  \includegraphics[width=0.8\textwidth]{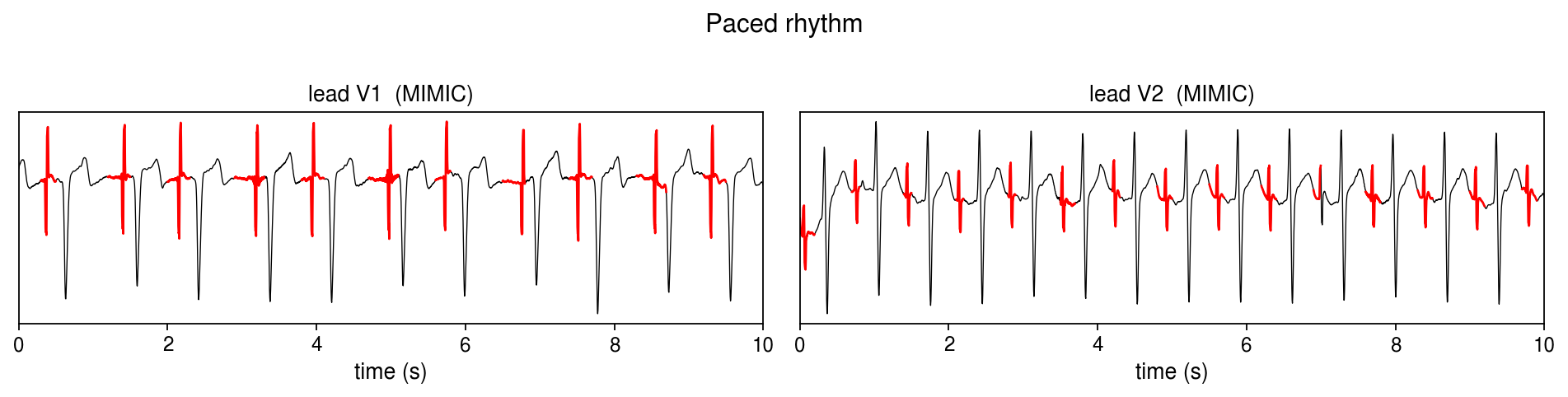}
  \caption{\textbf{Paced-rhythm localization.} Atom L6/5276 fires, shown red,
  on the pacing-spike artifact before each QRS complex, across two example
  ECGs (leads V1 and V2).}
  \label{fig:paced}

  \vspace{8pt}
  \includegraphics[width=0.8\textwidth]{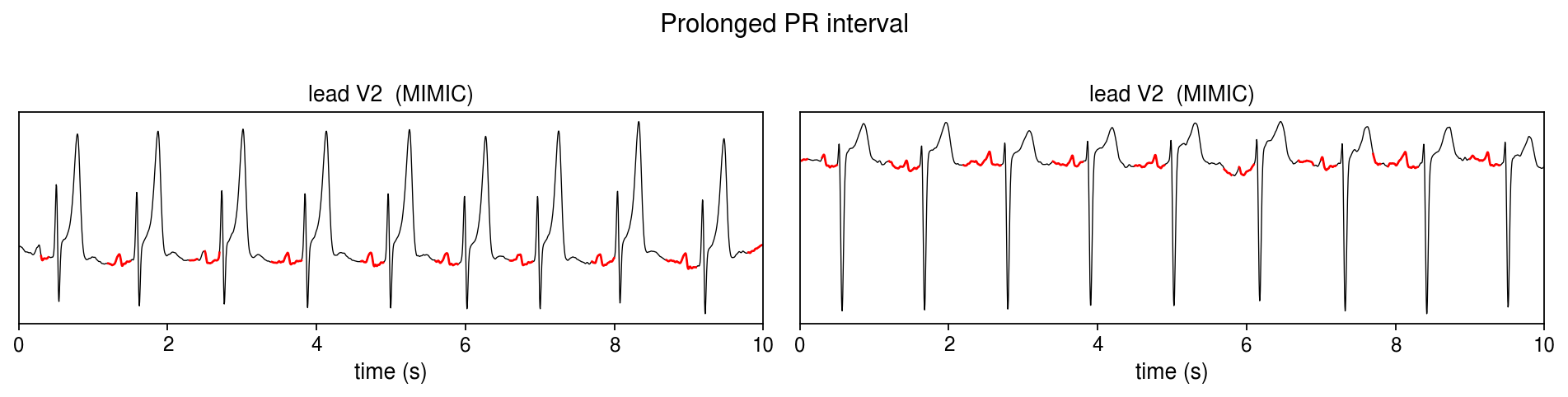}
  \caption{\textbf{Prolonged-PR localization.} Atom L6/4213 fires, shown red,
  on the extended PR segment between the P wave and QRS across two example
  ECGs (lead V2).}
  \label{fig:pr}
\end{figure*}

\figref{fig:rhythm} shows the rhythm-level atom referenced in
\secref{sec:localize}. A sinus-bradycardia atom (L6/3082) activates on the
\emph{elongated diastolic baseline
between beats}---the prolonged TP/RR interval---firing on every inter-beat
segment in lead~II and staying silent through the QRS--T complexes. Its
``feature'' is thus a multi-beat timing property that no single time point
causes, complementing the phenotype- and morphology-level atoms of
\figref{fig:layers}.

A second rhythm example is atrial flutter: atom L6/418 fires on the regular
sawtooth flutter waves in the baseline between QRS complexes
(\figref{fig:flutter}).

A paced-rhythm atom localizes a different waveform primitive: atom L6/5276
fires on the sharp pacing-spike artifact preceding each QRS complex
(\figref{fig:paced}).

An interval atom localizes the conduction delay itself: atom L6/4213 fires on
the prolonged PR segment between the P wave and the QRS
(\figref{fig:pr}).

\begin{figure*}[tp]
  \centering
  \includegraphics[width=0.8\textwidth]{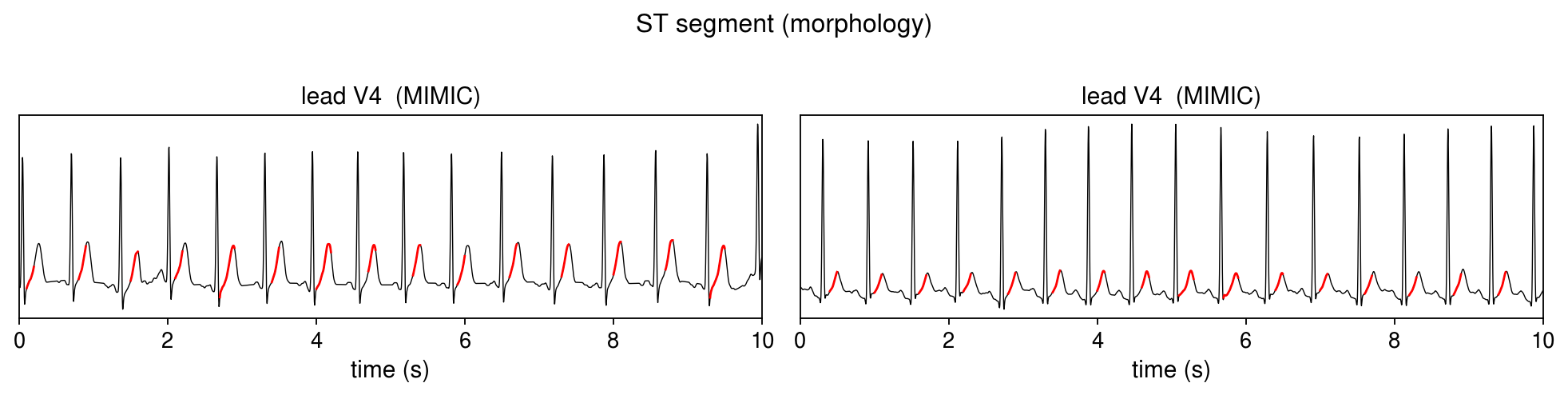}
  \caption{\textbf{Morphology-level localization: ST segment.} A morphology
  atom (L6/1305) fires, shown red, on the ST--T segment immediately after each
  QRS complex, across two MIMIC example ECGs (lead~V4).}
  \label{fig:st}
\end{figure*}

A morphology atom localizes a sub-beat waveform primitive: atom L6/1305 fires
on the ST--T segment immediately after each QRS complex, across two example
ECGs (\figref{fig:st}).

\subsection{Phenotype relationships: AUROC-fingerprint variant}
The phenotype-relationship structure of \figref{fig:relations} is robust to
the similarity metric. \figref{fig:relations_auroc} shows the same
curated concepts under a signed-AUROC fingerprint (each concept's vector of
signed per-atom AUROCs), whose cosine recovers the identical
conduction / wide-QRS block; per-pair values agree with the prototype
cosine.

\subsection{External validation on PTB-XL}
\label{app:ptbxl}
\begin{figure*}[tp]
  \centering
  \begin{minipage}[t]{0.49\textwidth}
    \centering
    \includegraphics[width=\linewidth]{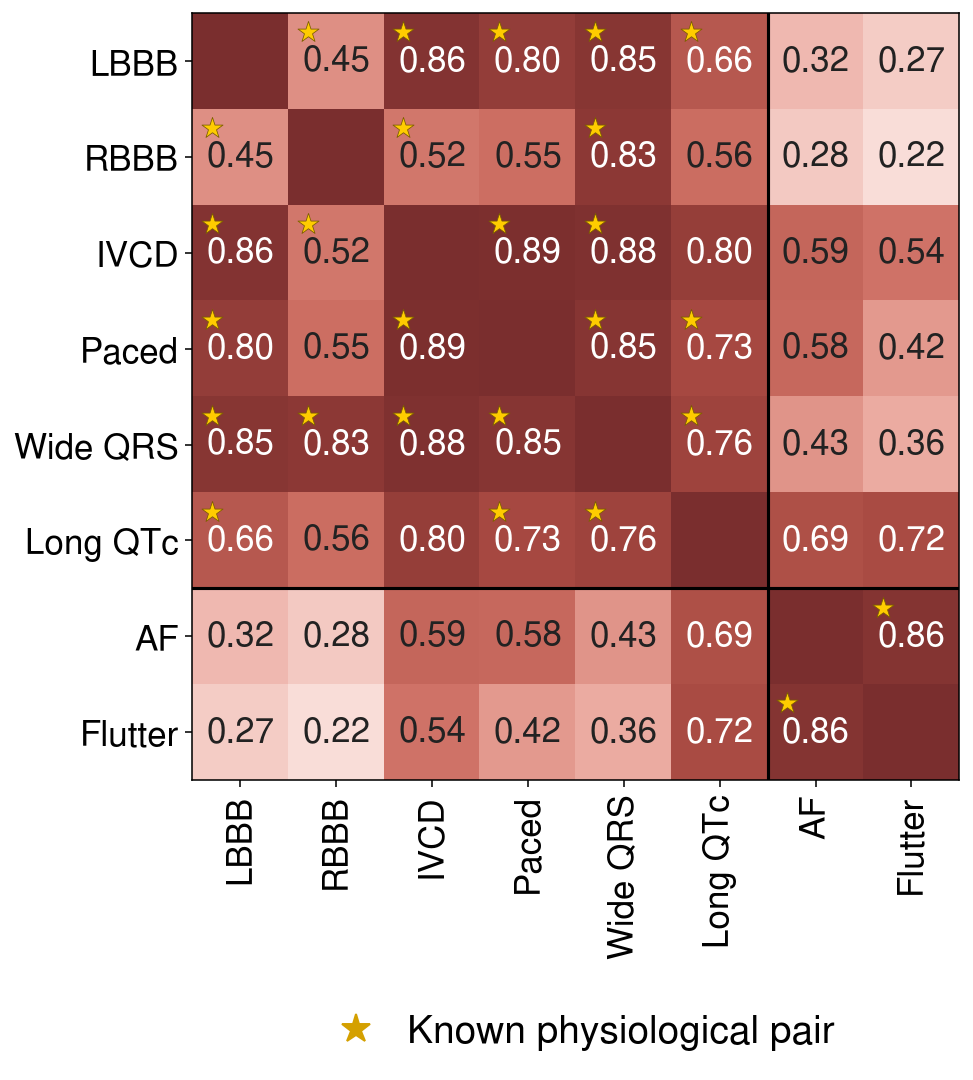}
    \caption{\textbf{Phenotype relationships under the AUROC-fingerprint
    metric}, cf.\ \figref{fig:relations}. Same curated concept set; cosine of
    signed-AUROC fingerprints. The data-driven block and star-marked known
    pairs are unchanged.}
    \label{fig:relations_auroc}
  \end{minipage}\hfill
  \begin{minipage}[t]{0.49\textwidth}
    \centering
    \includegraphics[width=\linewidth]{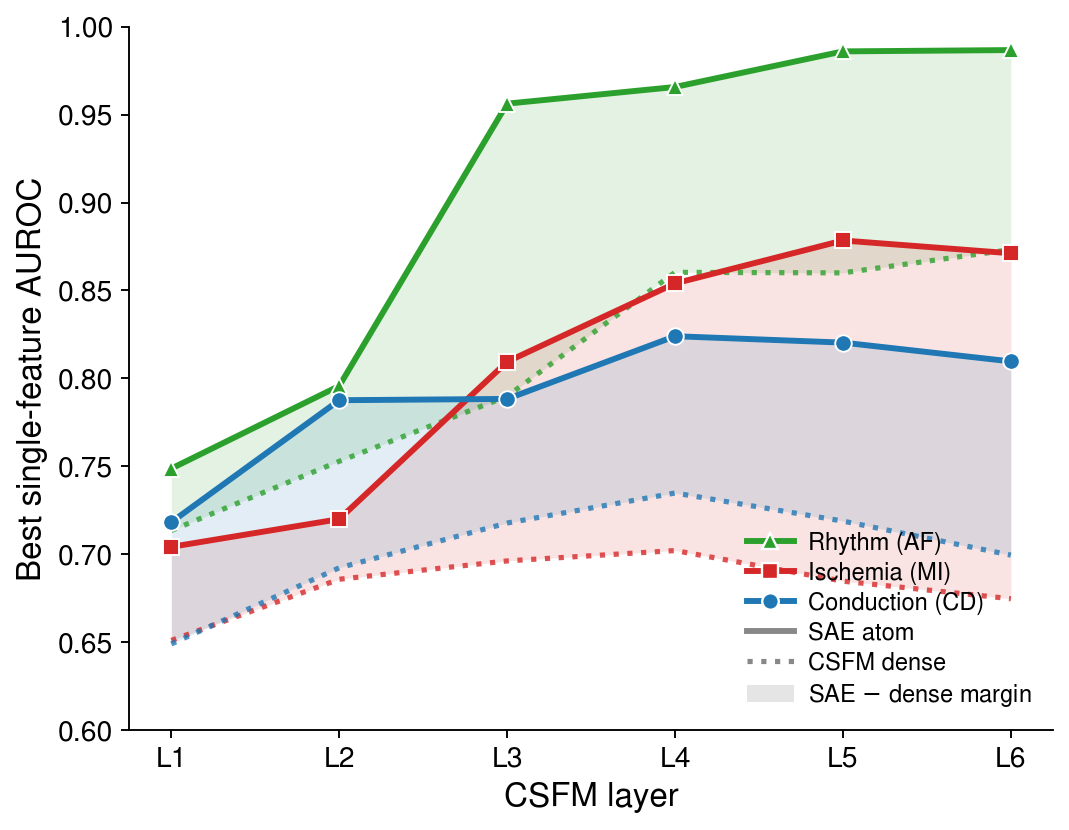}
    \caption{\textbf{External validation on PTB-XL: best-atom vs.\ best-dense
    across depth, diagnostic tasks.} Best single-feature test AUROC for three
    representative cardiologist-labeled diagnoses---rhythm/AF, ischemia/MI,
    conduction/CD; solid SAE atoms vs.\ dotted CSFM dense dimensions, shaded
    SAE-over-dense margin. Frozen MIMIC-trained SAEs applied to PTB-XL.}
    \label{fig:ptbxl_diag}
  \end{minipage}

  \vspace{4pt}
  \begin{minipage}[t]{0.49\textwidth}
    \centering
    \includegraphics[width=\linewidth]{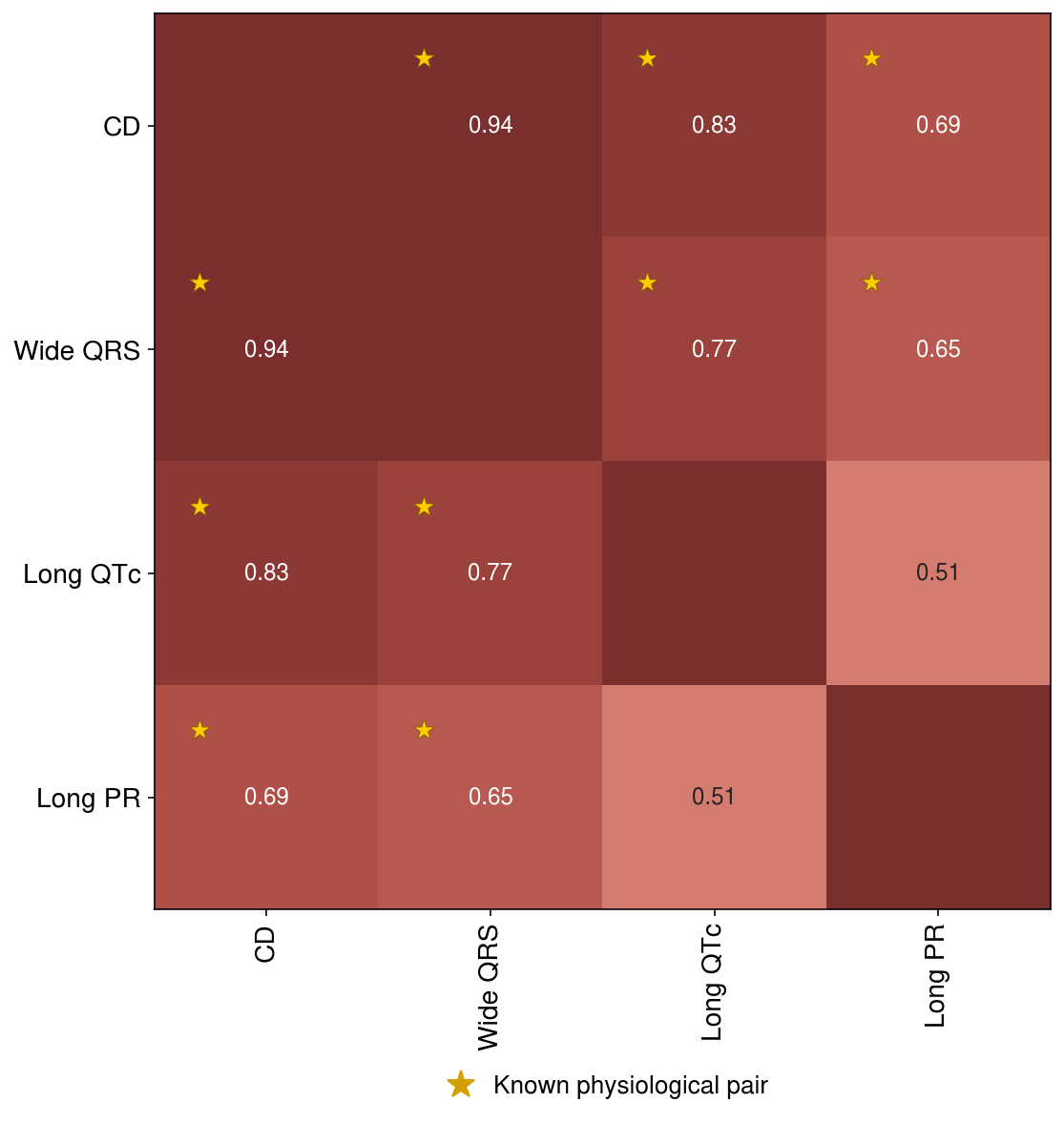}
    \caption{\textbf{External validation on PTB-XL: concept relationships in
    atom space.} Cosine similarity of PTB-XL AUROC-fingerprints for the
    conduction/wide-QRS concepts; yellow stars mark known cardiology pairs. The
    same physiological block recovered on MIMIC (\figref{fig:relations}) is
    reproduced on the external cohort.}
    \label{fig:ptbxl_cosine}
  \end{minipage}\hfill
  \begin{minipage}[t]{0.49\textwidth}
    \centering
    \includegraphics[width=\linewidth]{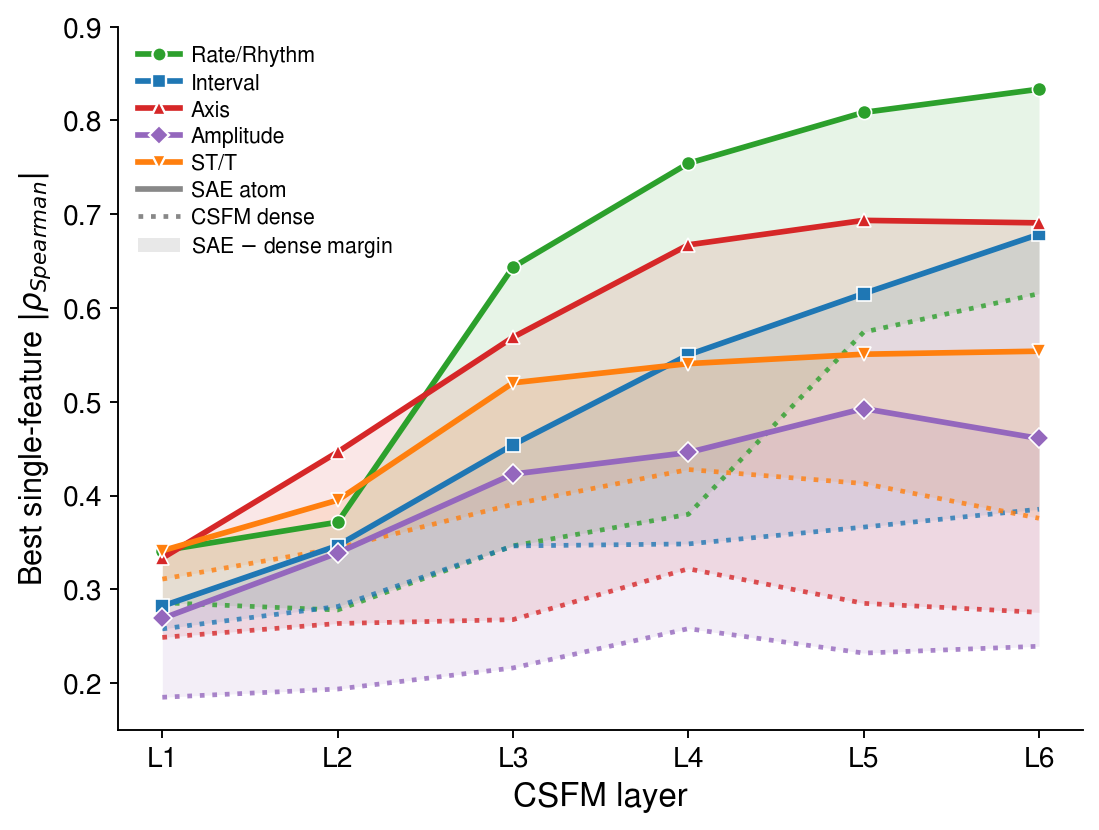}
    \caption{\textbf{External validation on PTB-XL: best-atom vs.\ best-dense
    across depth, 49 measurement concepts.} Best single-feature association
    $|\rho_{\mathrm{Spearman}}|$ per CSFM layer, averaged within each of the
    five concept families; solid SAE atoms vs.\ dotted CSFM dense, shaded
    margin. Atoms lead dense in every family at every layer.}
    \label{fig:ptbxl_concepts}
  \end{minipage}
\end{figure*}

To test whether the dictionary generalizes beyond its training distribution,
we apply the \emph{frozen}, MIMIC-trained per-layer SAEs---without any
retraining or re-estimation of their input statistics---to
PTB-XL~\cite{ptbxl}, an independent, cardiologist-labeled 12-lead cohort from
a different country and acquisition system ($21{,}799$ ECGs; patient-level
train/test split). PTB-XL signals are preprocessed identically to
pre-training and passed through CSFM and the SAEs to obtain per-ECG atom
activations; the atom-activation scale matches the MIMIC distribution
(median max $\approx 22.8$ vs.\ $24$), confirming the frozen dictionary is
in-distribution. We reproduce the best-single-feature analysis of
\figref{fig:layers}a against two independent label sources
(\figref{fig:ptbxl_diag}, \figref{fig:ptbxl_concepts}): the five cardiologist-
adjudicated diagnostic superclasses (AUROC) and $49$ automated
12SL/ECGDeli \emph{measurement} concepts grouped into five families
(association $|\rho_{\mathrm{Spearman}}|$). In both, the best SAE atom exceeds
the best dense dimension at \emph{every} layer with the margin widening with
depth---mean best-atom AUROC $0.825$ vs.\ dense $0.720$ over the diagnostic
tasks, and atoms ahead of dense in all five measurement families---so the
Layer-6 monosemanticity advantage of \figref{fig:layers}a is not an artifact
of the training cohort. Of the $49$ measurement concepts, only $\sim$5 (heart
rate, PR, QRS duration, QT/QTc, axis) have a counterpart in our MIMIC concept
set; the remaining $\sim$40 (amplitudes, areas, regional ST--T, additional
axes) are recovered by the frozen dictionary despite never being evaluated
during development.

Beyond per-concept detection, the atom-space geometry also transfers: on
PTB-XL, the cosine similarity of concept fingerprints reproduces the
conduction/wide-QRS block of \figref{fig:relations}
(\figref{fig:ptbxl_cosine}), recovering the interval-containment link
wide-QRS\,--\,long-QTc ($0.77$) and the conduction--QRS-widening relations
without any relational supervision.

\subsection{Automated atom descriptions: worked examples and the LLM prompt}
\label{app:autodesc}

\paragraph{Atom inclusion.}
Atoms are auto-interpreted only if their evidence can be constructed and
fairly validated; using pre-hoc activation/geometry statistics (not the
validation score), we exclude atoms that are \emph{dead} (negligible maximum
activation), \emph{rare} (fewer than $20$ high-activation ECGs, so a
strong/medium/weak example set cannot be built), \emph{near-constant} (firing
on $>\!90\%$ of ECGs, leaving no negatives), \emph{weak} (maximum activation
below a floor), or \emph{lead-diffuse} (per-lead activation entropy $>0.85$,
i.e.\ no localizable lead). The last criterion is motivated by a pilot: the
atoms whose descriptions failed validation were almost all lead-diffuse. Of
the $8192$ atoms, $2991$ ($\sim\!46\%$ of active atoms) pass; we interpret a
random sample of them.

\paragraph{Describe.}
For each included atom we take its highest-activating tokens, render the
corresponding lead$\times$beat-phase waveform crops with the activated region
highlighted plus a 12-lead context, and prompt the model with the images and
an activation-geometry summary (lead and beat-phase histograms; the same
geometry gives each atom a fixed, label-independent name, \appref{app:naming})
for a structured, geometry-first JSON description (dominant lead, beat phase,
waveform shape). No machine-report text is provided to the describer; an
ablation adding the full report text of top-activating ECGs did not change
faithfulness (median Spearman $\rho$ $0.665\!\to\!0.670$), ruling out
report-label leakage.

\paragraph{Validate.}
We validate each description by activation prediction. From the description
\emph{alone} (no activation overlay, no activation values), the model scores a
held-out montage of crops spanning the atom's \emph{high}, \emph{median}, and
\emph{low} activation levels; we correlate predicted with measured activation
across these graded buckets. We report the Spearman correlation $\rho$ because
the model's scores are ordinal and activations are heavily skewed. A
dissimilar-atom control (scoring a different concept's atom with this
description) gives a null near zero.

\begin{figure*}[tp]
  \centering
  \includegraphics[width=\textwidth]{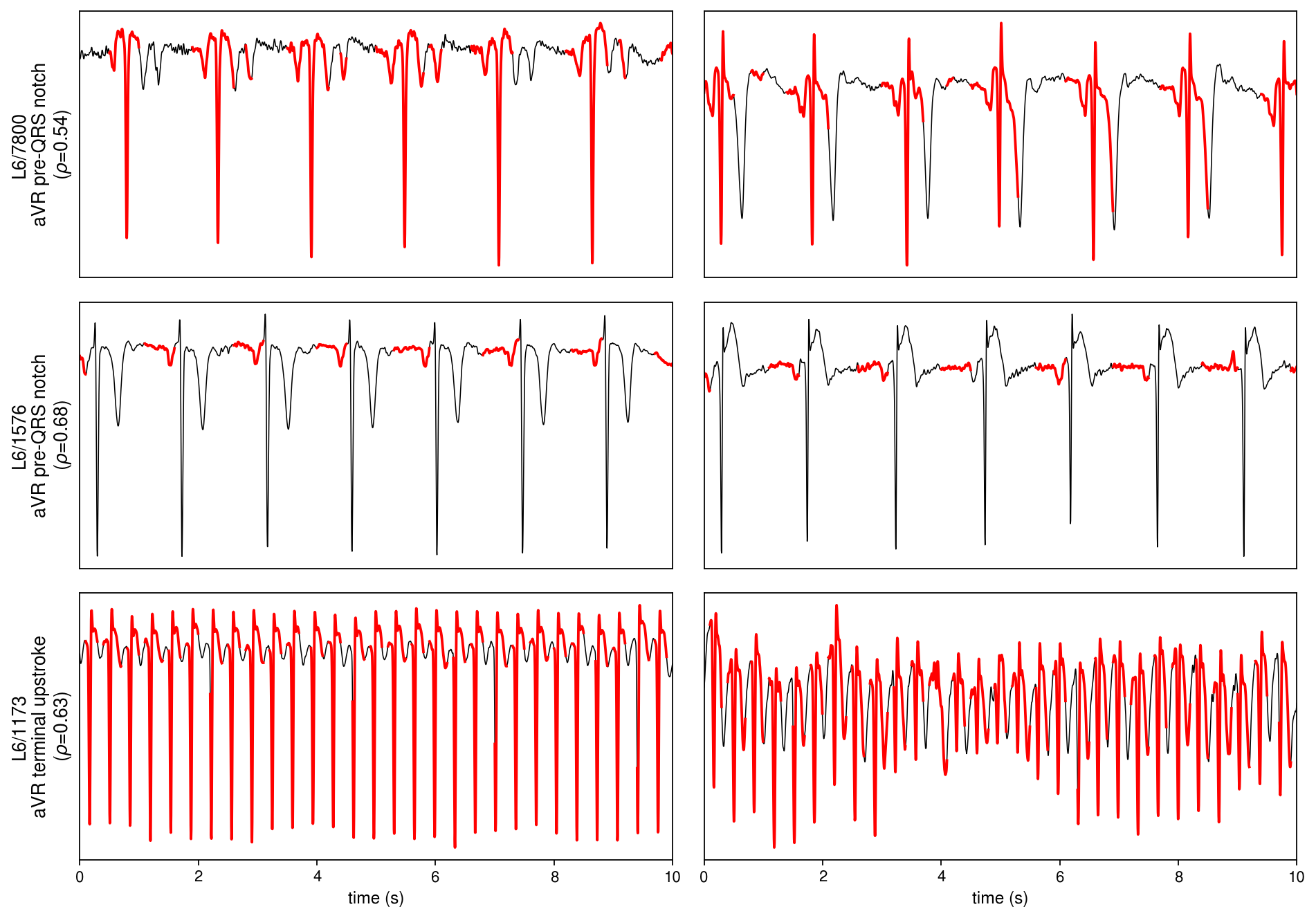}
  \caption{\textbf{Three example atoms described by the pipeline.} Each row is one
  atom (left label: its \texttt{gpt-5.4-mini} name and held-out faithfulness);
  the two panels are its top-activating ECGs (distinct records) in the atom's
  dominant lead, with red marking where the atom fires. All three localize to
  lead aVR at a specific beat phase: L6/7800, ``aVR pre-QRS notch'' (P--PR,
  Spearman $\rho{=}0.54$); L6/1576, ``aVR pre-QRS notch'' (P--PR, $\rho{=}0.68$);
  and L6/1173, ``aVR terminal upstroke'' (QRS-terminal, $\rho{=}0.63$).
  Faithfulness is the Spearman correlation between the model's held-out predicted
  and measured activations.}
  \label{fig:atom_desc_examples}
\end{figure*}

This section makes the automated description pipeline of \secref{sec:llm}
(\figref{fig:llm}) concrete on three Layer-6 atoms. Each description is produced
by \texttt{gpt-5.4-mini} from activation \emph{geometry} alone: the lead- and
beat-phase histograms of the atom's top-activating tokens, plus two evidence
images (eight local activation crops and a 12-lead context with the activated
patch highlighted). Clinical report text is passed only as auxiliary context and
the model is instructed to name atoms by geometry, not diagnosis. The description
is then validated by having the same model score held-out crops from the
description alone; we report the Spearman correlation $\rho$ between predicted and
measured activation. \figref{fig:atom_desc_examples} shows the three atoms with
their returned descriptions and faithfulness, and \figref{fig:prompt} reproduces
the exact prompt verbatim.

The verbatim \texttt{morphology\_description} field returned for each atom:
\begin{itemize}[leftmargin=1.4em,itemsep=2pt,topsep=2pt]
  \item \textbf{L6/1173:} ``In lead aVR, the atom fires on a consistent terminal
  QRS/early ST feature: a small upward rebound or positive notch immediately after
  a predominantly negative QRS complex, sometimes extending into the earliest ST
  segment.''
  \item \textbf{L6/1576:} ``A small pre-QRS deflection in aVR during the P/PR
  interval, typically a subtle notch or bump on a relatively flat baseline just
  before the QRS onset; examples are fairly consistent and mostly beat-locked.''
  \item \textbf{L6/7800:} ``A small pre-QRS deflection in aVR, usually a brief
  negative P/PR-era notch or dip just before the sharp QRS complex; the pattern is
  consistently localized to aVR across examples.''
\end{itemize}

\begin{figure}[H]
\small
\textbf{System prompt.}
\begin{lstlisting}
You interpret a sparse-autoencoder (SAE) atom of an ECG model. The atom activates on specific LOCAL
lead x time-patch (beat-phase) waveform patterns, shown by red-highlighted patches in local crops and a
12-lead context. Identify the LOCAL morphology the atom captures: which lead(s), which beat phase
(P/PR/QRS-onset/QRS-peak/QRS-terminal/ST/T), what waveform shape, and whether examples are consistent.
Name it by activation GEOMETRY, not by diagnosis, unless the waveform clearly supports it.
Report text is AUXILIARY only. Output JSON only.
\end{lstlisting}
\textbf{User message} (per atom; two images attached).
\begin{lstlisting}
Atom L6/<id>.
Activation geometry summary (primary): lead histogram over top tokens = {...}, beat-phase histogram = {...},
  modal_lead = <lead>, lead_group = <group>.
Image [1]: 8 local crops at top-activating patches (red = activated patch).
Image [2]: 12-lead context of the top token (red = activated patch).
AUXILIARY ONLY (do not lead with this): report enrichment = {...}, best phenotype assoc = <pheno> (AUROC <x>).
<JSON schema, below>
\end{lstlisting}
\textbf{Required JSON schema} (appended to the user message).
\begin{lstlisting}
Return ONLY this JSON:
{
 "short_name": "<=6 words, geometry-based",
 "primary_geometry": {
   "lead_pattern":    "single-lead Vx | lateral | inferior | anteroseptal | global",
   "beat_phase":      "P | PR | QRS-onset | QRS-peak | QRS-terminal | ST | T | non-beat-locked",
   "temporal_extent": "point | wave-segment | broad",
   "activation_consistency": "beat-locked | non-beat-locked"
 },
 "morphology_description": "the local waveform shape where the atom fires (no diagnosis unless supported)",
 "report_enrichment": ["aux phrases or empty"],
 "confidence": "high | medium | low",
 "caveat": "report enrichment auxiliary; primary interpretation = lead-time activation geometry"
}
\end{lstlisting}
\caption{Verbatim prompt used by the automated atom-description pipeline
(\secref{sec:llm}): a fixed system prompt, a per-atom user message (with two
evidence images attached), and the required JSON output schema.}
\label{fig:prompt}
\end{figure}

\subsection{Concept inventory}
\label{app:concepts}
\tabref{tab:concepts} lists every clinical concept CADENCE is evaluated
against, grouped by clinical family. Diagnostic \emph{phenotypes} are derived
by keyword matching on the MIMIC-IV-ECG machine-report text; \emph{continuous
measurements} are read from the machine-measurement numeric fields;
\emph{derived interval concepts} are clinical thresholds on those
measurements (used to build the concept-fingerprint heatmaps of
\secref{sec:relations}); \emph{demographics and outcome} come from the
linked MIMIC-IV records; and \emph{morphology primitives} are not a fixed
list but an emergent class of sub-diagnostic atoms localized to a single lead
and beat phase. Prevalences are fractions of the $150$k training cohort;
because rare phenotypes were over-sampled during cohort construction
(\appref{app:cohort}), these are cohort---not population---rates. Two
report keywords (\texttt{t\_wave\_abnormality}, \texttt{nonspecific\_st\_t})
matched almost no records in this cohort and are excluded from downstream
phenotype analyses.

\begin{table*}[tp]
  \centering
  \begin{minipage}[t]{0.40\textwidth}
    \centering
    \caption{Phenotype and morphology features recovered by CADENCE, grouped by
    category.}
    \label{tab:features}
    \small
    \begin{tabular}{ll}
      \toprule
      Category & Feature \\
      \midrule
      \multicolumn{2}{l}{\textit{Phenotype features}} \\
      Rhythm \& ectopy    & sinus bradycardia \\
                          & atrial fibrillation \\
                          & atrial flutter \\
                          & premature ventricular complex \\
                          & paced rhythm \\
      Conduction \& axis  & left bundle branch block \\
                          & right bundle branch block \\
                          & left anterior fascicular block \\
                          & PR prolongation \\
                          & left-axis deviation \\
      Chamber \& voltage  & left ventricular hypertrophy \\
                          & low QRS voltage \\
      Ischemia \& infarct & inferior infarct \\
                          & anterior infarct \\
                          & ST depression \\
      Repolarization      & QT prolongation \\
      \midrule
      \multicolumn{2}{l}{\textit{Morphology primitives (lead $\times$ beat-phase)}} \\
      P / PR segment      & atrial / pre-QRS deflection \\
      QRS onset           & Q wave / initial deflection \\
      QRS peak            & R-wave peak \\
      QRS terminal        & terminal notch / R$'$ \\
      ST segment          & ST-segment deviation \\
      T wave              & T-wave repolarization \\
      \bottomrule
    \end{tabular}
  \end{minipage}\hfill
  \begin{minipage}[t]{0.56\textwidth}
    \centering
    \caption{Concept inventory used throughout CADENCE. Prevalence is the
    fraction of the 150k training cohort positive for each phenotype
    (over-sampled for rare classes; not population rates).}
    \label{tab:concepts}
    \small
    \begin{tabular}{lll}
      \toprule
      Concept & Definition / source & Prev.\ (\%) \\
      \midrule
      \multicolumn{3}{l}{\emph{Rhythm \& ectopy}} \\
      Sinus rhythm & report keyword & 49.0 \\
      Atrial fibrillation & report keyword & 10.9 \\
      Atrial flutter & report keyword & 4.5 \\
      Sinus bradycardia & report keyword & 16.7 \\
      Sinus tachycardia & report keyword & 13.5 \\
      Premature ventricular (PVC) & report keyword & 3.8 \\
      Premature atrial (PAC) & report keyword & 2.5 \\
      \midrule
      \multicolumn{3}{l}{\emph{Conduction \& axis}} \\
      LBBB & report keyword & 4.1 \\
      RBBB & report keyword & 6.8 \\
      First-degree AV block & report keyword & 1.9 \\
      Left-axis deviation & report keyword & 11.7 \\
      Right-axis deviation & report keyword & 2.4 \\
      \midrule
      \multicolumn{3}{l}{\emph{Chamber \& voltage}} \\
      LVH & report keyword & 10.5 \\
      Low QRS voltage & report keyword & 14.6 \\
      \midrule
      \multicolumn{3}{l}{\emph{Ischemia \& repolarization}} \\
      Myocardial infarction & report keyword & 29.6 \\
      ST depression & report keyword & 1.3 \\
      ST elevation & report keyword & 8.5 \\
      Prolonged QT & report keyword & 9.2 \\
      T-wave abnormality$^{\dagger}$ & report keyword & $<$0.1 \\
      Nonspecific ST-T$^{\dagger}$ & report keyword & $<$0.1 \\
      \midrule
      \multicolumn{3}{l}{\emph{Continuous measurements}} \\
      Heart rate (HR) & measurement (bpm) & --- \\
      PR interval & measurement (ms) & --- \\
      QRS duration & measurement (ms) & --- \\
      QT / QTc & measurement (ms) & --- \\
      Electrical axis & measurement ($^{\circ}$) & --- \\
      \midrule
      \multicolumn{3}{l}{\emph{Derived interval concepts}} \\
      Wide QRS & QRS $\geq$ 120\,ms & --- \\
      Long QTc & QTc $\geq$ 460\,ms & --- \\
      Long PR & PR $\geq$ 200\,ms & --- \\
      Bradycardia & HR $<$ 60\,bpm & --- \\
      Tachycardia & HR $>$ 100\,bpm & --- \\
      \midrule
      \multicolumn{3}{l}{\emph{Demographics \& outcome}} \\
      Age & linked record & --- \\
      Mortality & linked record & --- \\
      \midrule
      \multicolumn{3}{l}{\emph{Morphology primitives}} \\
      Lead $\times$ beat-phase atoms & emergent (P, PR, QRS-on/ & --- \\
      \quad(sub-diagnostic) & \quad peak/term, ST, T) & \\
      \bottomrule
    \end{tabular}\\[2pt]
    {\footnotesize $^{\dagger}$ near-zero prevalence in this cohort; excluded
    from phenotype analyses.}
  \end{minipage}
\end{table*}

\subsection{Model and training details}
\tabref{tab:backbone} summarizes the frozen CSFM backbone we interpret,
and \tabref{tab:sae} the sparse autoencoder we train on its Layer-6
activations. All models are implemented in PyTorch~\cite{pytorch2019}, and
downstream probes use scikit-learn~\cite{sklearn2011}.

\begin{table}[H]
  \centering
  \caption{Frozen ECG foundation-model backbone (CSFM). We use the released
  pretrained encoder as a fixed feature extractor; the pretraining decoders
  and task head are discarded.}
  \label{tab:backbone}
  \small
  \begin{tabular}{ll}
    \toprule
    Property & Value \\
    \midrule
    Model & CSFM (Cardiac Sensing FM), \texttt{Tiny} variant \\
    Modality & 12-lead ECG (multi-modal model: ECG/PPG/text) \\
    Pretraining & self-supervised masked reconstruction \\
    Pretraining scale & $\sim$1.7M individuals \\
    Hidden width $d$ & 768 \\
    Transformer layers & 6 \\
    Patch size & 25 samples (100\,ms) \\
    Tokens per ECG & $12~\text{leads}\times100~\text{patches}=1200$ \\
    Interpreted layer & 6 (residual stream) \\
    Preprocessing & resample $500\!\to\!250$\,Hz; per-channel \\
              & clean (neurokit2); per-lead $z$-norm \\
    Use in this work & frozen (no fine-tuning) \\
    \bottomrule
  \end{tabular}
\end{table}

\begin{table}[H]
  \centering
  \caption{Sparse autoencoder trained on frozen CSFM Layer-6 tokens.}
  \label{tab:sae}
  \small
  \begin{tabular}{ll}
    \toprule
    Hyperparameter & Value \\
    \midrule
    Type & BatchTopK SAE \\
    Input dim $d$ & 768 \\
    Dictionary size $F$ & 8192 \\
    Active codes $k$ (batch-avg $L_0$) & 128 \\
    Decoder & unit-normalized columns (per step) \\
    Input normalization & centered; scaled to $\mathbb{E}\lVert x\rVert^2\!\approx\!d$ \\
    Objective & reconstruction MSE $+$ AuxK \\
    AuxK ($k_{\mathrm{aux}}$, coef, dead thresh.) & 256, 0.03, 200 steps \\
    Optimizer & Adam~\cite{adam2015}, lr $4\times10^{-4}$ \\
    Batch size & 4096 tokens \\
    Training steps & 30000 \\
    Training tokens & $\sim\!9\times10^{6}$ (9M) Layer-6 tokens \\
    Explained variance & 0.968 \\
    Dead atoms & 1710 ($\sim\!21\%$) \\
    $F$ selection & grid search (\appref{app:grid}) \\
    \bottomrule
  \end{tabular}
\end{table}

\subsection{Training cohort construction}
\label{app:cohort}
The SAE is trained on Layer-6 tokens extracted from a phenotype-stratified,
subject-disjoint sample of MIMIC-IV-ECG~\cite{mimicivecg,mimiciv,physionet}. The cohort is built as follows.
\begin{enumerate}[leftmargin=1.4em,itemsep=1pt,topsep=2pt]
  \item \textbf{Candidate pool.} All $800{,}036$ records in the
  MIMIC-IV-ECG \texttt{record\_list.csv}.
  \item \textbf{Weak phenotype labels.} For every record we scan the
  \texttt{machine\_measurements} report text and assign $\sim$20 binary
  phenotype labels (e.g., AF, LBBB, PVC) by keyword matching. These labels
  are used only to \emph{balance sampling}, never as supervision for the SAE.
  \item \textbf{Subject-disjoint split.} Subjects are partitioned by a hash
  of the subject id ($\mathrm{hash}(\text{subject\_id})\bmod 10^4 < 1500
  \Rightarrow$ eval, else train), so no patient appears in both the training
  and evaluation sets and there is no subject leakage.
  \item \textbf{Stratified sampling to a target size.} Within the training
  pool we sample toward a target of $150{,}000$ ECGs: rarest phenotypes are
  drawn first (each capped at $12{,}000$ ECGs) to over-represent uncommon
  morphologies, at most $2$ ECGs are taken per subject, and the remainder is
  filled at random. The held-out evaluation set ($\sim$15k ECGs, used for all
  downstream analyses) is sampled identically from the disjoint eval pool.
\end{enumerate}
Each of the $150{,}000$ training ECGs contributes $60$ Layer-6 tokens
(evenly sampled across the 12 leads), yielding the $\sim$9M-token training
corpus. We chose $150{,}000$ ECGs rather than the full pool because the
dictionary saturates well below this scale: retraining on a $10\times$ larger
corpus ($9\text{M}\!\to\!98.8$M tokens) at the same $F{=}8192$, $k{=}128$
left explained variance essentially unchanged ($0.968\!\to\!0.970$) while
\emph{increasing} the dead-atom fraction ($21\%\!\to\!33\%$), indicating that
9M tokens already suffices for this dictionary capacity.

\subsection{Geometry-based atom naming}
\label{app:naming}
Each atom receives a \emph{label-independent} primary name of the form
``\{lead group\} \{beat phase\} atom,'' computed only from where the atom fires
on the waveform, so naming never depends on a diagnostic label. We profile an
atom on its top-8 activating test ECGs, giving a $12$-lead activation-mass
vector and an activation template aligned to neurokit-detected
R-peaks~\cite{neurokit2}. The \emph{lead group} is the single lead, anatomical
group, or ``global'' pattern that carries the activation mass; the \emph{beat
phase} is read from the template's peak offset relative to the
R-peak---P-wave, QRS-onset, QRS, QRS-terminal/J, ST, or T-wave---with wide
beat-locked templates named ``broad-QRS'' and non-beat-locked atoms named
``rhythm.'' Thus ``global broad-QRS'' denotes a wide, beat-locked
depolarization across most leads, the signature of paced or bundle-branch
widening; ``broad-QRS'' is a geometric proxy for, not a measurement of, QRS
duration. Separately, and only as an \emph{auxiliary} annotation, we report the
GE-MUSE report statement most enriched among each atom's top-activating ECGs;
this shares its source with the diagnostic labels and is used for description
only, the geometric name being the atom's identity. The fixed thresholds are as
follows.

We profile each atom on its top-8 activating test ECGs; each patch spans
$25$ samples ($100$\,ms). \textbf{Lead group}, from the per-lead activation
mass $m$: a single lead if $m_{\max}/\sum m > 0.5$; else an anatomical group
(inferior, lateral, anteroseptal, \dots) if that group holds more than half
the mass; else ``global'' if at least nine of the twelve leads are jointly
active (each exceeding $0.4\,m_{\max}$); else ``\{group\}-predominant.''
\textbf{Beat phase}, relative to neurokit-detected R-peaks: an atom is
``non-beat-locked rhythm'' if more than half of its active patches fall
outside a per-beat window $[R\!-\!2,R\!+\!6]$. Otherwise we average the
activation over offsets $d\in[-4,+8]$ patches into an R-aligned template,
whose peak offset $d^\star$ and width (patches above half the template peak)
give the phase: width $\ge 3$ patches ($\gtrsim\!300$\,ms) is ``broad-QRS'';
else P-wave ($d^\star\!\le\!-2$), QRS-onset ($-1$), QRS-locked ($0$),
QRS-terminal/J ($+1$), ST-segment ($+2$), or T-wave (later).

\end{document}